\documentclass[sigconf]{acmart}
\usepackage[export]{adjustbox}

\AtBeginDocument{%
  \providecommand\BibTeX{{%
    \normalfont B\kern-0.5em{\scshape i\kern-0.25em b}\kern-0.8em\TeX}}}

\setcopyright{acmcopyright}
\copyrightyear{2022}
\acmYear{2022}
\acmDOI{XXXXXXX.XXXXXXX}

\acmConference[FDG '22]{FDG '22: International Conference on the Foundations of Digital Games}{September 5--8, 2022}{Athens, Greece}
\acmBooktitle{FDG '22: International Conference on the Foundations of Digital Games, September 5--8, 2022, Athens, Greece}
\acmPrice{15.00}
\acmISBN{978-1-4503-XXXX-X/18/06}

\usepackage{graphicx}
\usepackage{caption}
\usepackage{subcaption}
\usepackage{hyperref}

\usepackage{bm}

\usepackage{tikz}
\usetikzlibrary{positioning}
\usetikzlibrary{matrix}

\usepackage{pgfplots, pgfplotstable}
\pgfplotsset{compat=newest}

\pgfplotsset{
    show sum on top/.style={
        /pgfplots/scatter/@post marker code/.append code={%
            \node[
                at={(normalized axis cs:%
                        \pgfkeysvalueof{/data point/x},%
                        \pgfkeysvalueof{/data point/y})%
                },
                anchor=south,
            ]
            {\pgfmathprintnumber{\pgfkeysvalueof{/data point/y}}};
        },
    },
}

\begin{document}

% Usar o termo 'illuminating the space' é comum em títulos de trabalhos que usam MAP-Elites, por isso escolhi
% Illuminating the Space of Dungeons Maps and Locked-door Missions Through MAP-Elites
\title{Illuminating the Space of Dungeon Maps, Locked-door Missions and Enemy Placement Through MAP-Elites}

\author{Breno M. F. Viana}
\orcid{1234-5678-9012}
\affiliation{%
 \institution{Universidade de São Paulo}
 \city{São Carlos}
 \state{São Paulo}
 \country{Brazil}
 \postcode{13566-590}
}
\email{bmfviana@gmail.com}

\author{Leonardo T. Pereira}
\orcid{1234-5678-9012}
\affiliation{%
 \institution{Universidade de São Paulo}
 \city{São Carlos}
 \state{São Paulo}
 \country{Brazil}
 \postcode{13566-590}
}
\email{leonardop@usp.br}

\author{Claudio F. M. Toledo}
\orcid{1234-5678-9012}
\affiliation{%
 \institution{Universidade de São Paulo}
 \city{São Carlos}
 \state{São Paulo}
 \country{Brazil}
 \postcode{13566-590}
}
\email{claudio@icmc.usp.br}

\begin{abstract} % 0.3pgs
Procedural Content Generation (PCG) methods are valuable tools to speed up the game development process. Moreover, PCG may also be present in games as features, such as the procedural dungeon generation (PDG) in \textit{Moonlighter} (Digital Sun, 2018). This paper introduces an extended version of an evolutionary dungeon generator by incorporating a MAP-Elites population. Our dungeon levels are discretized with rooms that may have locked-door missions and enemies within them. We encoded the dungeons through a tree structure to ensure the feasibility of missions. We performed computational and user feedback experiments to evaluate our PDG approach. They show that our approach accurately converges almost the whole MAP-Elite population for most executions. Finally, players' feedback indicates that they enjoyed the generated levels, and they could not indicate an algorithm as a level generator.
\end{abstract}




\maketitle

\section{Introduction} % 0.7pgs

Creating game content from scratch is a hardworking task for game designers.
Thus, game developers may apply Procedural Content Generation (PCG) to speed up the development process \cite{togelius2016introduction}.
Some of the many examples of successful games which apply PCG techniques are \textit{No Man's Sky} by Hello Games, and \textit{Moonlighter} by Digital Sun \cite{ref:nomanssky, moonlighter}.
The latter applies PCG to generate its dungeon levels, a.k.a. Procedural Dungeon Generation (PDG), which is popular both in the research community and game industry \cite{ref:viana2021procedural}.

According to van der Linden et al., dungeons are labyrinth environments composed of challenges, rewards, and puzzles interrelated with the playspace and time, where the former is the physical layout where the game takes place \cite{ref:van2014procedural, ref:dormans2011generating}.
Furthermore, these games present missions, i.e., a set of goals that the player must accomplish \cite{ref:pereira2021procedural}.
Examples of such a mission are: defeat/kill some enemies, talk to characters, collect some items, collect keys to open locked doors, among others.
Locked-door missions are considered puzzles interrelated with the playspace \cite{ref:de2019procedural, ref:viana2021procedural}.

The present paper approaches the locked-door missions for dungeons, applying PCG by extending the evolutionary algorithm (EA) introduced in Pereira et al. \cite{ref:pereira2021procedural}. The previous work presented an EA capable of generating dungeon levels with locked-door missions that match the game designer parameters for levels and missions. Our extended version offers two main contributions from \cite{ref:pereira2021procedural} and PDG related works.
The first one is to advance from the previous EA by evolving also the enemies distribution through the levels' rooms. The second contribution is the application of a MAP-Elites algorithm for enhancing Quality Diversity (QD) in content generation, taking into account the level design with lock, keys, and enemies placement.
QD algorithms are very relevant for PCG purposes since they can create different content in a single run, according to Gravina et al. \cite{ref:gravina2019procedural}.

Following the definition of game facets in \cite{liapis2019maestro}, our approach fits in orchestrating of levels and narrative (as lock and key missions), totaling two creative facets orchestrated concurrently by a single algorithm. Our population has two dimensions and is mapped regarding leniency, as defined by Smith et al. \cite{ref:smith2018graph}, and exploration coefficient, similar to the concept defined by Liapis et al. \cite{ref:liapis2013towards}. The evolutionary parameters are closer to those used by Pereira et al. \cite{ref:pereira2021procedural}; however, we added the number of enemies as a parameter, and we replaced the number of generations by time-limit as a stop criterion to ensure most levels converge. Also, besides generating levels to match the entered characteristics, our approach aims to balance the distribution of enemies in the levels and provide more diversity based on leniency and exploration criteria as previously mentioned.

The results show that our algorithm accurately converges all dungeon levels on the MAP-Elites defined for most executions.
Furthermore, we evaluated our levels by collecting volunteers' feedback after playing a game prototype with our levels, and the main feedback was that most of them enjoyed the gameplay.
% Coloco isso no resumo?
Besides, most players could not point out if an algorithm created the levels.

We structured the paper as follows: \autoref{sec:relatedwork} presents the related works; \autoref{sec:methodology} describes the representation of our dungeon levels and our evolutionary level generation approach;
\autoref{sec:results} presents and discusses the results of our experiments;
finally, \autoref{sec:conclusion} presents the conclusions and future works.

\section{Related Works} \label{sec:relatedwork} % 1pg

Most works on PDG apply search-based approaches in their solutions \cite{ref:viana2021procedural}.
Gravina et al. reported that QD approaches are relevant for PCG purposes since they can generate a variety of contents in a single execution without losing quality \cite{ref:gravina2019procedural}.
Within this class of algorithms, there are the Illumination Algorithms that return sets of the best-found solutions, which are discretized in a map regarding their features \cite{ref:mouret2015illuminating}.
In PCG research, the illumination through MAP-Elites-based approaches has been increasing \cite{ref:gravina2019procedural, ref:viana2021procedural}.
We present some related PDG works, in which some of them applied MAP-Elites approaches. \autoref{tab:relatedworks} summarizes the comparison of our paper with related works reviewed through this section.

\begin{table}[!htb]
\caption{Summarizing of related works on dungeon generation and comparison with this work.}
\label{tab:relatedworks}
\begin{tabular}{@{}ccccc@{}}
\toprule
Work & Content & Enemy & Missions & MAP-Elites \\
\midrule
\cite{ref:alvarez2019empowering} & Room & - & - & $\checkmark$ \\
\cite{ref:charity2020mech} & Room & $\checkmark$ & - & $\checkmark$ \\
\cite{ref:dormans2010adventures} & Level & - & $\checkmark$ & - \\
\cite{ref:dormans2011level} & Level & - & $\checkmark$ & - \\
\cite{ref:van2013designing} & Level & $\checkmark$ & $\checkmark$ & - \\
\cite{ref:gellel2020hybrid} & Level & - & $\checkmark$ & - \\
\cite{ref:pereira2021procedural} & Level & - & $\checkmark$ & - \\
\cite{ref:liapis2013towards} & Level & $\checkmark$ & - & - \\
\midrule
Our work & Level & $\checkmark$ & $\checkmark$ & $\checkmark$ \\
\bottomrule
\end{tabular}
\end{table}

% Generation through MAP-Elites

Alvarez et al. extended the Evolutionary Dungeon Design (EDD), a mixed-initiative tool introduced by Baldwin et al., to provide several suggestions of changes presented to the user as a matrix of rooms \cite{ref:alvarez2019empowering, ref:baldwin2017mixed}.
This feature was possible due to the Interactive Constrained MAP-Elites (CME), which they introduced.
This method can generate suggestions based on the room during the edition process.
Each user modification leads to the generation of new suggestions, and it is possible to map the matrix of suggestions into linearity, symmetry, and other specific metrics of the EDD.

Charity et al. introduced an automatic method based on CME to generate rooms for general games based on their mechanics \cite{ref:charity2020mech}.
They applied this approach in four different games with different game mechanics sets by mapping the CME's matrix according to their mechanics.
They also used the General Video Game Artificial Intelligence (GVG-AI) framework to generate the initial population and the framework's agents in the fitness functions.
The fitness functions are based on the survival and conclusion time of the agents.
They claim that the generated rooms can be used as game tutorials to teach the players how to use the game mechanics.

Both previous works applied MAP-Elites in their algorithms to generate rooms \cite{ref:alvarez2019empowering, ref:charity2020mech}.
We also apply such an approach, but we generate levels instead of rooms.
Consequently, we based our MAP-Elites' feature descriptors on different metrics.

% Mission generation algorithms

Following, we describe works that tackled the problem of generating levels with missions.
Dormans generated dungeons through a Generative Grammar (GG) \cite{ref:dormans2010adventures}.
The approach creates a graph of missions and applies it to generate the play-space for Action-Adventure games.
Later, Dormans used the previous GG and a model-driven approach in a mixed-initiative tool to generate level sketches \cite{ref:dormans2011level}.
The introduced model-driven approach ``evolves'' the dungeons through model transformations.
Similar to Dormans, van der Linden et al. presented a GG-based method for level generation \cite{ref:van2013designing}.
The work uses gameplay as a vocabulary to control the generative process.
The graphs' nodes express player actions as gameplay design constraints, and they are described semantically, e.g., ``fight melee enemy'' and ``pickup health potion''.
Thus, the gameplay grammar allows designers to specify their expected gameplay. This work, like ours, also presents the placement of enemies.

Gellel and Sweetser combined Dormans'  generative grammar with a non-traditional Cellular Automata (CA) \cite{ref:gellel2020hybrid, ref:dormans2010adventures}.
First, the GG creates a string that codifies the locked-door missions and the level rooms.
After that, they applied the new CA method based on random neighbor selection to generate the play-space.
This method is composed of two rules that place the rooms for each character of the mission string.

So far, the works that generate levels with locked-door missions have applied GG approaches to their solutions.
Differently from them, Pereira et al. presented a search-based algorithm for dungeon level generation with locked-door missions for Action-Adven\-ture games \cite{ref:pereira2021procedural}.
The proposed algorithm is the base for the approach introduced in this paper.
The missions' goal is to collect keys to open locked doors in the levels and find a symbol, similar to Zelda's triforce.
A tree structure represents the dungeon levels to ensure feasibility, and the Genetic Programming approach evolves the levels and missions.
The keys are in the trees' nodes (rooms), and the doors are in the trees' edges (corridors).
We chose to enhance this algorithm in our research, as it creates feasible dungeons, accurate to the designer's needs, fast, with no need for training, and very few numerical inputs.
Our approach goes beyond it by introducing the enemy element in the generation and ensuring QD.

% Generation with enemies

Works like those of van der Linden, Alvarez et al., and Charity perform enemy placement in their approaches \cite{ref:van2013designing, ref:alvarez2019empowering, ref:charity2020mech}.
Besides them, Liapis introduced a search-based approach where such property is more important than the previous works.
The author presented an approach for dungeon generation composed of two stages applying a FI2Pop GA.
The first stage generates dungeon sketches by strategically placing eight segments representing dungeon rooms, which describe impassable (wall), passable rooms and define the number of enemies and rewards in each room.
In the second step, each segment becomes a room.
Each room evolves independently to create a cavern environment, following connections between the segments and their types.
For instance, the enemies are strategically placed around a reward \cite{ref:liapis2017multi}.

As for the orchestration of content, in Liapis et al. \cite{liapis2019maestro} review on the topic, none of the presented works that applied the orchestration algorithm also used QD.
The same holds for more recent papers focused on the topic of orchestration that we found, like the algorithm of Karavolo et al. \cite{karavolos2018facets} for first-person shooter content orchestration or the experiment of Prager et al. on the effects of the combination of visuals and audio facets \cite{prager2019facets}. 
Thus, we emphasize our contribution in creating and orchestrating two creative facets (level and, partially, narrative) with a single algorithm and enforcing QD on them. We found that this is the first time that a single algorithm created such facets focusing on QD.

\section{Methodology} \label{sec:methodology} % 2pgs

This section describes the representation of our dungeon levels and the MAP-Elites approach we applied to evolve them.

\subsection{Representation}

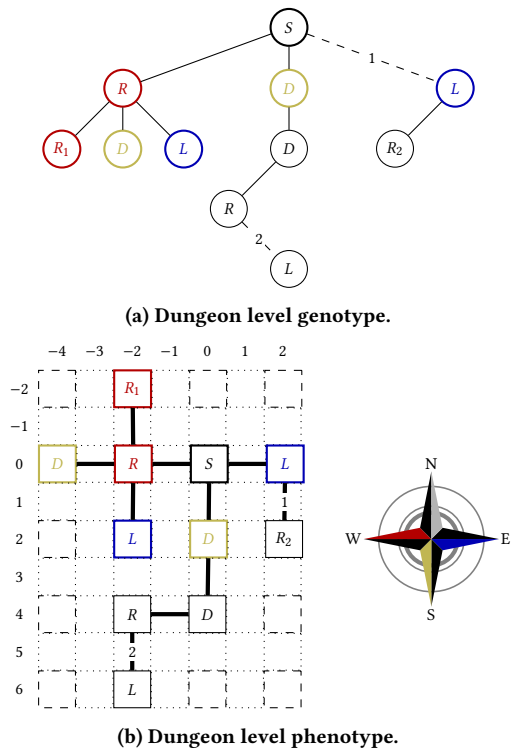
\begin{figure}[!t]
    \centering
    \begin{subfigure}[t]{\columnwidth}
        \centering
        \begin{tikzpicture}[every node/.style={scale=0.7}]
        % Node style
        \tikzstyle{node} = [draw, circle, minimum size=0.7cm]

        % Depth 0, root
        \node[node, thick] (n1) at (0,0) {$S$};
    
        % Depth 1
        \node[node, thick, yellow!70!black] (n2) [below = 0.3cm of n1] {$D$};
        \node[node, thick, blue!70!black] (n3) [right = 1.7cm of n2] {$L$};
        \node[node, thick, red!70!black] (n4) [left = 1.7cm of n2] {$R$};
    
        % Depth 2
        \node[node, thick, yellow!70!black] (n5) [below = 0.3cm of n4] {$D$};
        \node[node, thick, blue!70!black] (n6) [right = 0.3cm of n5] {$L$};
        \node[node, thick, red!70!black] (n7) [left = 0.3cm of n5] {$R_1$};
        
        \node[node] (n8) [below = 0.3cm of n2] {$D$};
        
        \node[node, opacity=0] (an3) [below = 0.3cm of n3] {$0$};
        \node[node] (n9) [left = 0.3cm of an3] {$R_2$};
        
        % Depth 3
        \node[node, opacity=0] (an8) [below = 0.3cm of n8] {$0$};
        \node[node] (n10) [left = 0.3cm of an8] {$R$};
        
        % Depth 3
        \node[node, opacity=0] (an9) [below = 0.3cm of n10] {$0$};
        \node[node] (n11) [right = 0.3cm of an9] {$L$};
        
        % Connect nodes
        \path[draw] (n1) -- (n2);
        \path[draw, dashed] (n1) -- (n3)
            node[pos=0.5,fill=white,inner sep=1]{1};
        \path[draw] (n1) -- (n4);
        
        \path[draw] (n4) -- (n5);
        \path[draw] (n4) -- (n6);
        \path[draw] (n4) -- (n7);
        
        \path[draw] (n2) -- (n8);
        \path[draw] (n8) -- (n10);
        \path[draw, dashed] (n10) -- (n11)
            node[pos=0.5,fill=white,inner sep=1]{2};
        
        \path[draw] (n3) -- (n9);
        \end{tikzpicture}
        \subcaption{Dungeon level genotype.}
        \label{fig:tree-encoding}
    \end{subfigure}
    
    \begin{subfigure}[t]{\columnwidth}
        \centering
        \begin{tikzpicture}[every node/.style={scale=0.7}]
        % Node style
        \tikzstyle{room} = [draw, minimum size=0.7cm]
        \tikzstyle{corridor} = [draw, minimum size=0.7cm, dotted]
        \tikzstyle{empty_room} = [draw, minimum size=0.7cm, dashed]
        \tikzstyle{none} = [draw, minimum size=0.7cm, opacity=0]
        \tikzstyle{coordinate} = [minimum size=0.7cm]
        
        \node[coordinate]
            (l0n0) [] { };
        \node[coordinate]
            (l0n1) [right = 0cm of l0n0] {$-4$};
        \node[coordinate]
            (l0n2) [right = 0cm of l0n1] {$-3$};
        \node[coordinate]
            (l0n3) [right = 0cm of l0n2] {$-2$};
        \node[coordinate]
            (l0n4) [right = 0cm of l0n3] {$-1$};
        \node[coordinate]
            (l0n5) [right = 0cm of l0n4] {$0$};
        \node[coordinate]
            (l0n6) [right = 0cm of l0n5] {$1$};
        \node[coordinate]
            (l0n7) [right = 0cm of l0n6] {$2$};
        \node[coordinate]
            (l1n0) [below = 0cm of l0n0] {$-2$};
        \node[empty_room]
            (l1n1) [right = 0cm of l1n0] { };
        \node[corridor]
            (l1n2) [right = 0cm of l1n1] { };
        \node[room, thick, red!70!black]
            (l1n3) [right = 0cm of l1n2] {$R_1$};
        \node[corridor]
            (l1n4) [right = 0cm of l1n3] { };
        \node[empty_room]
            (l1n5) [right = 0cm of l1n4] { };
        \node[corridor]
            (l1n6) [right = 0cm of l1n5] { };
        \node[empty_room]
            (l1n7) [right = 0cm of l1n6] { };
        \node[coordinate]
            (l2n0) [below = 0cm of l1n0] {$-1$};
        \node[corridor]
            (l2n1) [right = 0cm of l2n0] { };
        \node[none]
            (l2n2) [right = 0cm of l2n1] { };
        \node[corridor]
            (l2n3) [right = 0cm of l2n2] { };
        \node[none]
            (l2n4) [right = 0cm of l2n3] { };
        \node[corridor]
            (l2n5) [right = 0cm of l2n4] { };
        \node[none]
            (l2n6) [right = 0cm of l2n5] { };
        \node[corridor]
            (l2n7) [right = 0cm of l2n6] { };
        \node[coordinate]
            (l3n0) [below = 0cm of l2n0] {$0$};
        \node[room, thick, yellow!70!black]
            (l3n1) [right = 0cm of l3n0] {$D$};
        \node[corridor]
            (l3n2) [right = 0cm of l3n1] { };
        \node[room, thick, red!70!black]
            (l3n3) [right = 0cm of l3n2] {$R$};
        \node[corridor]
            (l3n4) [right = 0cm of l3n3] { };
        \node[room, thick]
            (l3n5) [right = 0cm of l3n4] {$S$};
        \node[corridor]
            (l3n6) [right = 0cm of l3n5] { };
        \node[room, thick, blue!70!black]
            (l3n7) [right = 0cm of l3n6] {$L$};
        \node[coordinate]
            (l4n0) [below = 0cm of l3n0] {$1$};
        \node[corridor]
            (l4n1) [right = 0cm of l4n0] { };
        \node[none]
            (l4n2) [right = 0cm of l4n1] { };
        \node[corridor]
            (l4n3) [right = 0cm of l4n2] { };
        \node[none]
            (l4n4) [right = 0cm of l4n3] { };
        \node[corridor]
            (l4n5) [right = 0cm of l4n4] { };
        \node[none]
            (l4n6) [right = 0cm of l4n5] { };
        \node[corridor]
            (l4n7) [right = 0cm of l4n6] { };
        \node[coordinate]
            (l5n0) [below = 0cm of l4n0] {$2$};
        \node[empty_room]
            (l5n1) [right = 0cm of l5n0] { };
        \node[corridor]
            (l5n2) [right = 0cm of l5n1] { };
        \node[room, thick, blue!70!black]
            (l5n3) [right = 0cm of l5n2] {$L$};
        \node[corridor]
            (l5n4) [right = 0cm of l5n3] { };
        \node[room, thick, yellow!70!black]
            (l5n5) [right = 0cm of l5n4] {$D$};
        \node[corridor]
            (l5n6) [right = 0cm of l5n5] { };
        \node[room]
            (l5n7) [right = 0cm of l5n6] {$R_2$};
        \node[coordinate]
            (l6n0) [below = 0cm of l5n0] {$3$};
        \node[corridor]
            (l6n1) [right = 0cm of l6n0] { };
        \node[none]
            (l6n2) [right = 0cm of l6n1] { };
        \node[corridor]
            (l6n3) [right = 0cm of l6n2] { };
        \node[none]
            (l6n4) [right = 0cm of l6n3] { };
        \node[corridor]
            (l6n5) [right = 0cm of l6n4] { };
        \node[none]
            (l6n6) [right = 0cm of l6n5] { };
        \node[corridor]
            (l6n7) [right = 0cm of l6n6] { };
        \node[coordinate]
            (l7n0) [below = 0cm of l6n0] {$4$};
        \node[empty_room]
            (l7n1) [right = 0cm of l7n0] { };
        \node[corridor]
            (l7n2) [right = 0cm of l7n1] { };
        \node[room]
            (l7n3) [right = 0cm of l7n2] {$R$};
        \node[corridor]
            (l7n4) [right = 0cm of l7n3] { };
        \node[room]
            (l7n5) [right = 0cm of l7n4] {$D$};
        \node[corridor]
            (l7n6) [right = 0cm of l7n5] { };
        \node[empty_room]
            (l7n7) [right = 0cm of l7n6] { };
        \node[coordinate]
            (l8n0) [below = 0cm of l7n0] {$5$};
        \node[corridor]
            (l8n1) [right = 0cm of l8n0] { };
        \node[none]
            (l8n2) [right = 0cm of l8n1] { };
        \node[corridor]
            (l8n3) [right = 0cm of l8n2] { };
        \node[none]
            (l8n4) [right = 0cm of l8n3] { };
        \node[corridor]
            (l8n5) [right = 0cm of l8n4] { };
        \node[none]
            (l8n6) [right = 0cm of l8n5] { };
        \node[corridor]
            (l8n7) [right = 0cm of l8n6] { };
        \node[coordinate]
            (l9n0) [below = 0cm of l8n0] {$6$};
        \node[empty_room]
            (l9n1) [right = 0cm of l9n0] { };
        \node[corridor]
            (l9n2) [right = 0cm of l9n1] { };
        \node[room]
            (l9n3) [right = 0cm of l9n2] {$L$};
        \node[corridor]
            (l9n4) [right = 0cm of l9n3] { };
        \node[empty_room]
            (l9n5) [right = 0cm of l9n4] { };
        \node[corridor]
            (l9n6) [right = 0cm of l9n5] { };
        \node[empty_room]
            (l9n7) [right = 0cm of l9n6] { };
        
        % Corridors
        \path[draw, ultra thick] (l3n5) -- (l3n3);
        \path[draw, ultra thick] (l3n3) -- (l5n3);
        \path[draw, ultra thick] (l3n3) -- (l3n1);
        \path[draw, ultra thick] (l3n3) -- (l1n3);
        
        \path[draw, ultra thick] (l3n5) -- (l3n7);
        \path[draw, ultra thick] (l3n7) -- (l5n7)
            node[pos=0.5,fill=white,inner sep=2]{1};
        
        \path[draw, ultra thick] (l3n5) -- (l5n5);
        \path[draw, ultra thick] (l5n5) -- (l7n5);
        \path[draw, ultra thick] (l7n5) -- (l7n3);
        \path[draw, ultra thick] (l7n3) -- (l9n3)
            node[pos=0.5,fill=white,inner sep=2]{2};

        % Wind rose
        \node[draw, circle, semithick, white!50!black, minimum size=2cm] (wind_rose) [right = 1cm of l5n7] {};
        \node[draw, circle, ultra thick, white!50!black, minimum size=1cm] (wr1) at (wind_rose.center) {};
        \node[draw, circle, semithick, white!50!black, minimum size=1.25cm] (wr2) at (wind_rose.center) {};
        % East
        \fill[black] (wind_rose.center) -- ++(0.15, 0.15) -- ++(0.75, -0.15);
        \fill[blue!70!black] (wind_rose.center) -- ++(0.15, -0.15) -- ++(0.75, 0.15);
        % South
        \fill[black] (wind_rose.center) -- ++(0.15, -0.15) -- ++(-0.15, -0.75);
        \fill[yellow!70!black] (wind_rose.center) -- ++(-0.15, -0.15) -- ++(0.15, -0.75);
        % West
        \fill[black] (wind_rose.center) -- ++(-0.15, -0.15) -- ++(-0.75, 0.15);
        \fill[red!70!black] (wind_rose.center) -- ++(-0.15, 0.15) -- ++(-0.75, -0.15);
        % North
        \fill[black] (wind_rose.center) -- ++(-0.15, 0.15) -- ++(0.15, 0.75);
        \fill[white!70!black] (wind_rose.center) -- ++(0.15, 0.15) -- ++(-0.15, 0.75);

        \node (N) [above = 0.85cm of wind_rose.center] {N};
        \node (W) [left = 0.85cm of wind_rose.center] {W};
        \node (S) [below = 0.85cm of wind_rose.center] {S};
        \node (E) [right = 0.85cm of wind_rose.center] {E};
        \end{tikzpicture}
        \subcaption{Dungeon level phenotype.}
        \label{fig:grid-translation}
    \end{subfigure}
    
    \caption{Handcrafted example of a genotype-phenotype level translation. (a) presents the level genotype and (b) presents the resulting phenotype. The root node $\boldsymbol{S}$ represents the starting room. Nodes with $\boldsymbol{R}$ (Right), $\boldsymbol{D}$ (Down), and $\boldsymbol{L}$ (Left) represent the direction the parent node connects with them. The numbers in the nodes are keys. The numbers in the dashed edges are the locks. Rooms are always placed in even values of the x and y coordinates, while corridors are placed in coordinates with different parities. By comparing the node colors with the wind rose, we see that a parent room is always considered in the north direction regarding any of its child rooms.}
    \label{fig:tree-representation}
\end{figure}

One individual in our evolutionary algorithm states a dungeon level, where a tree structure represents such an individual, as shown in \autoref{fig:tree-encoding}.
Each node defines a room that encodes its type and position in the dungeon.
We have a Key Room ($KR$), which indicates an available key to open a locked door, a Locked Room ($LR$) with a locked door, and a Normal Room ($NR$) that has nothing special.
The node also encodes the number of enemies in that room and the room position concerning the parent node: Right ($R$), Down ($D$), and Left ($L$).
We place the room assuming that its parent room (node) is in the north, positioning it ($R$, $D$, and $L$) correctly when decoding the individual into a dungeon level.

%The 2D position of the rooms regards the room position in a 2D grid, where the starting room is always in position $(0, 0)$. 
%The dungeon's remaining rooms are placed around the starting room according to the directions of its child rooms and connected via corridors, with parent rooms always at North ($N$) regarding its children. \autoref{fig:tree-representation} examples a grid translated from a tree representation.
To ensure no room overlaps the level, we decode the tree representation (genotype) to a 2D grid (phenotype).
If there are overlapping rooms, the branch that causes the overlap is removed from the tree.
Thus, we ensure that no level is infeasible following the process of branch removal detailed in \cite{ref:pereira2021procedural}.
A single key can open a locked door; therefore, the keys are bound with their locks through a shared ID.
Rooms can have only a key, only a locker, or none of them; they can never have only one of each or multiples of a kind at the same time.
Moreover, our representation does not require keys to be collected and unlocked in a specific sequence.

%A pair of IDs for locks and keys defines the locked-door missions, e.g., the node $R_2$ in \autoref{fig:tree-encoding} shows a $KR$ with the key ID number 1 that will open the corridor represented by the arc $(R,L)$ with a number 2 over it, where $R$ is a $LR$. 

\subsection{Generation Process}

% Breno: Seria bom colocar essa informação? Não?
% Leo: Eu acho que sim. Falar quais os inputs é bem importante e acho que vão reclamar de não estar no artigo.
Our dungeon generation process evolves tree structures of feasible levels.
The parameters that our algorithm receives are the number of rooms, number of keys, number of locks, number of enemies, and linear coefficient (linearity).
%They define the goal characteristics that the generated levels should have.
%Therefore, as the approach of Pereira et al. \cite{ref:pereira2021procedural}, we aim to minimize the distance between the generated levels' attributes with the entered parameters.
We designed our approach to evolve dungeons by preserving diversity and optimizing quality.
To do so, we applied a MAP-Elites approach for variety by mapping the feature descriptors (or dimensions), weighting the leniency of enemies and exploration coefficient.
To measure the leniency of enemies in our levels, we apply the \autoref{eq:leniency} presented by Smith et al. \cite{ref:smith2018graph}.
The leniency is calculated by the number of safe rooms, i.e., without enemies, divided by the total number of rooms.
\begin{equation} \label{eq:leniency}
    D_{\text{leniency}} =  \frac{\mathit{Number~of~Safe~Rooms}}{\mathit{Total~Rooms}}
\end{equation}
\autoref{eq:exploration_coefficient} measures our exploration coefficient, inspired by the exploration measure introduced by Liapis et al. \cite{ref:liapis2013towards}.
We run a flood fill algorithm between rooms to simulate the map coverage, where the reached rooms represent the required exploration from each starting room and its corresponding goal room.
\begin{equation} \label{eq:exploration_coefficient}
    D_{\text{exploration}} = \frac{1}{\#RR} \sum_{(r_s, r_g) \in RR} \frac{\mathit{Coverage}(r_s, r_g)}{\mathit{Total~Rooms}}
\end{equation}
where $RR$ is the set of pairs of reference rooms containing the pair of starting and goal rooms and all pairs of key and locked rooms; $\#RR$ is the size of $RR$; $r_s$ is the room where the flood fill starts, and; $r_g$ is the goal room where the algorithm ends.

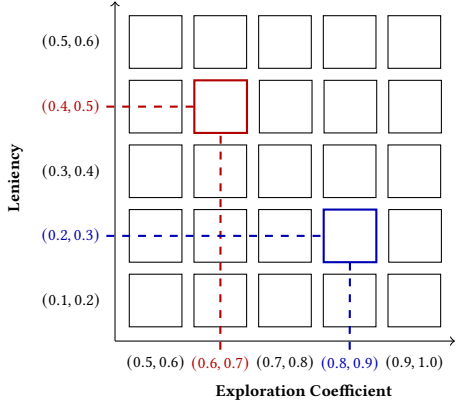
\begin{figure}[!t]
    \centering
    \begin{tikzpicture}[every node/.style={scale=0.7}]
    % Node style
    \tikzstyle{node} = [draw, minimum size=1cm]

    \node[node] (l1c1) at (0,0) { };
    \node[node] (l1c2) [right = 0.15cm of l1c1] { };
    \node[node] (l1c3) [right = 0.15cm of l1c2] { };
    \node[node] (l1c4) [right = 0.15cm of l1c3] { };
    \node[node] (l1c5) [right = 0.15cm of l1c4] { };
    \node[node] (l2c1) [below = 0.15cm of l1c1] { };
    \node[node, thick, red!70!black] (l2c2) [right = 0.15cm of l2c1] { };
    \node[node] (l2c3) [right = 0.15cm of l2c2] { };
    \node[node] (l2c4) [right = 0.15cm of l2c3] { };
    \node[node] (l2c5) [right = 0.15cm of l2c4] { };
    \node[node] (l3c1) [below = 0.15cm of l2c1] { };
    \node[node] (l3c2) [right = 0.15cm of l3c1] { };
    \node[node] (l3c3) [right = 0.15cm of l3c2] { };
    \node[node] (l3c4) [right = 0.15cm of l3c3] { };
    \node[node] (l3c5) [right = 0.15cm of l3c4] { };
    \node[node] (l4c1) [below = 0.15cm of l3c1] { };
    \node[node] (l4c2) [right = 0.15cm of l4c1] { };
    \node[node] (l4c3) [right = 0.15cm of l4c2] { };
    \node[node, thick, blue!70!black] (l4c4) [right = 0.15cm of l4c3] { };
    \node[node] (l4c5) [right = 0.15cm of l4c4] { };
    \node[node] (l5c1) [below = 0.15cm of l4c1] { };
    \node[node] (l5c2) [right = 0.15cm of l5c1] { };
    \node[node] (l5c3) [right = 0.15cm of l5c2] { };
    \node[node] (l5c4) [right = 0.15cm of l5c3] { };
    \node[node] (l5c5) [right = 0.15cm of l5c4] { };
    
    % Axis
    \node (ax) [below left = 0.15cm of l5c1] { };
    \node (ax_y1) [above left = 0.15cm of l1c1] { };
    \node (ax_y2) [below right = 0.15cm of l5c5] { };

    \draw[->] (ax.center) -- (ax_y1.center);
    \path (ax) -- node [rotate=90,below left=-0.45cm and 1.5cm] {\textbf{Leniency}} (ax_y1);
    \node (le1) [left = 0.3cm of l1c1] {$(0.5,0.6)$};
    \node[thick, red!70!black] (le2) [left = 0.3cm of l2c1] {$(0.4,0.5)$};
    \node (le3) [left = 0.3cm of l3c1] {$(0.3,0.4)$};
    \node[thick, blue!70!black] (le4) [left = 0.3cm of l4c1] {$(0.2,0.3)$};
    \node (le5) [left = 0.3cm of l5c1] {$(0.1,0.2)$};

    \draw[->] (ax.center) -- (ax_y2.center);
    \path (ax) -- node [below right=0.5cm and -1cm] {\textbf{Exploration Coefficient}} (ax_y2);
    \node (ce1) [below = 0.3cm of l5c1] {$(0.5,0.6)$};
    \node[thick, red!70!black] (ce2) [below = 0.3cm of l5c2] {$(0.6,0.7)$};
    \node (ce3) [below = 0.3cm of l5c3] {$(0.7,0.8)$};
    \node[thick, blue!70!black] (ce4) [below = 0.3cm of l5c4] {$(0.8,0.9)$};
    \node (ce5) [below = 0.3cm of l5c5] {$(0.9,1.0)$};

    \draw[dashed, thick, red!70!black] (le2) -- (l2c2);
    \draw[dashed, thick, red!70!black] (ce2) -- (l2c2);
    
    \draw[dashed, thick, blue!70!black] (le4) -- (l4c4);
    \draw[dashed, thick, blue!70!black] (ce4) -- (l4c4);
    \end{tikzpicture}

    \caption{The map of MAP-Elites population. The red cell represents a dungeon with leniency between 0.4 and 0.5 and an exploration coefficient between 0.6 and 0.7. The blue cell represents a dungeon with leniency between 0.2 and 0.3 and an exploration coefficient between 0.8 and 0.9. Thus, the blue level has more reference rooms further to each other than the red one, and it also has more rooms with enemies.}
    \label{fig:map-representation}
\end{figure}

Since our equations result in values between 0 and 1, we discretized such dimensions.
For the leniency dimension, the intervals are (0.5, 0.6), (0.4, 0.5), (0.3, 0.4), (0.2, 0.3), and (0.2, 0.1).
Levels with greater leniency values have most rooms without enemies or some of them with several enemies.
For exploration coefficient, the intervals are (0.5, 0.6), (0.6, 0.7), (0.7, 0.8), (0.8, 0.9), and (1.0, 0.9).
Levels with exploration coefficient values lesser than these lead to rooms much closer to each other.
\autoref{fig:map-representation} presents our approach's map.

The proposed MAP-Elites application will map 25 individuals based on the defined intervals. When the map receives a new individual, we must calculate the feature descriptors to place it in the correct entry of the MAP-Elites table. If an individual fills a map cell and a new one hits the same cell, the latter replaces the former if it has a better fitness; otherwise, we discard the new individual.

The evolutionary process starts generating individuals for the initial population by following the initialization algorithm described in their work to create rooms, keys, and lockers. However, we introduced two changes in the initialization procedure. First, we are dealing with the placement of enemies in our approach; therefore, after generating each dungeon, we place enemies in random rooms, one by one, except by the starting and goal rooms.
% BRENO, vc estava falando de inimigos, agora passou a falar de fechadura, pq? O trecho a seguir ficou desconexo e também está confuso. Explique nos comentários, por favor:
% Existe uma chance de o nível gerado aleatoriamente não tenha nenhuma fechadura, que é como definimos a sala objetivo/final. Então eu adicionei pra poder calcular o coeficiente de exploração do nível.
Besides, if the created level has no locker, we add one with a key to ensure that we can calculate the goal room.
If this level also has keys, we first remove one of them.
% Second, we add individuals to the initial population until it reaches 20 individuals.
% Essa quantidade é parâmetro do evolutivo também, então passei esse 20 pro primeiro parágrafo dos resultados.
Second, we add individuals to the initial population until it reaches $n$ individuals.
Since the initialized individuals may hit the same entry in the MAP-Elites table, it can take a while. In this case, the population size does not change once the best individual is always kept for that entry.

Next, we evolve the population using the time-limit stopping criterion.
Pereira et al. create an intermediate population that always replaces the current one, except by the best individual found so far \cite{ref:pereira2021procedural}.
In our case, after stating the intermediate population, we try to insert its individuals in the MAP-Elites population. 
Our intermediate population has new individuals created from two parents, which are chosen using tournament selection with two competitors.

The crossover randomly selects one parent node as the cut point to swap the selected nodes. After the swap, we remove all overlaps and rebuild the grid. We try to preserve the locks and keys of the original branches in the new individual by applying the repair algorithm described in \cite{ref:pereira2021procedural}.
%, but the repair stops if one of the swapped branches cannot store all the missions (i.e., $KR$ and $LR$ nodes) of the other one. In this case,  another pair of cut points is selected. If no feasible pair of cut points are found, the crossover cannot be performed, and none individuals are returned.
%Given that our map is small, with  a total capacity of 25 individuals, and the crossover may fail in generating new individuals, we always perform it.

We always apply crossover, while mutation has a chance of 15\% to be applied, where a pair of a lock and key has 50\% chance to be added or removed from the tree structure. We visit the tree structure through a breadth-first order to add a pair and convert a random $NR$ into a $KR$.
Next, we do the same to convert a random $NR$ into a $LR$ among the non-visited rooms. To remove a pair, we randomly select a $KR$, and its related $LR$, converting both into $NR$ nodes.
% \textcolor{red}{ Breno, veja se está ok assim,senão complemente ou explique nos comentários: If there are enemies, they are sent to another empty room randomly chosen.}
% Essa descrição não é o suficiente, segue a descrição mais completa:
% Depois da etapa de adição/remoção de missões começa a etapa de transferência de inimigos. Duas salas são escolhidas para transferir e outra para receber. Contudo, existe a possibilidade da seleção de salas escolherem duas salas sem inimigos nenhum, então não há transferência. Se a sala que transfere não tem inimigos e a que recebe tem, essa última vai transferir. A quantidade é escolhida de forma aleatória, mas tem que ser pelo menos 1 inimigo.
After this, we perform an enemy transfer operation.
To do so, we select two rooms to transfer and to receive them.
If both rooms have no enemies, nothing is done.
If the receiver has enemies and the transferer does not have them, we swap the rooms.
Then, we randomly chose from 1 to the transferer's number of enemies to move to the receiver room.

\begin{table*}[ht]
    \centering
    \caption{Results of fitness obtained after 30 executions of our approach. Each table caption represents a set of parameters: (number of rooms)-(number of keys)-(number of locks)-(number of enemies)-(linear coefficient). Each table cell corresponds to an Elite. Descriptors for leniency values: L1 = (0.5,0.6), L2 = (0.4,0.5), L3 = (0.3,0.4), L4 = (0.2,0.3), L5 = (0.1,0.2). Descriptors for values of exploration coefficient: E1 = (0.5,0.6), E2 = (0.6,0.7), E3 = (0.7,0.8), E4 = (0.8,0.9), E5 = (0.9,1.0).}
    \label{tab:fitness_tables}
    
    \begin{subtable}[t]{.5\textwidth}
    \centering
    \caption{15-3-2-20-2.}
    \label{tab:sp1}
    \begin{tabular}{cccccc}
    \toprule
   &              E1 &              E2 &              E3 &              E4 &              E5 \\
    \midrule
L1 &   1.02$\pm$0.61 &   0.84$\pm$0.40 &   0.83$\pm$0.40 &   0.85$\pm$0.38 &  1.06$\pm$0.59 \\
L2 &   0.33$\pm$0.47 &   0.25$\pm$0.40 &   0.23$\pm$0.39 &   0.26$\pm$0.38 &  0.49$\pm$0.57 \\
L3 &   0.09$\pm$0.46 &   0.02$\pm$0.40 &   0.00$\pm$0.40 &   0.04$\pm$0.39 &  0.26$\pm$0.57 \\
L4 &  -0.12$\pm$0.46 &  -0.21$\pm$0.39 &  -0.22$\pm$0.39 &  -0.18$\pm$0.39 &  0.03$\pm$0.54 \\
L5 &   0.05$\pm$0.82 &  -0.15$\pm$0.63 &  -0.19$\pm$0.60 &  -0.15$\pm$0.62 &  0.30$\pm$0.93 \\
    \bottomrule
    \end{tabular}
    \end{subtable}
    ~
    \begin{subtable}[t]{.5\textwidth}
    \centering
    \caption{20-4-4-30-1.}
    \label{tab:sp2}
    \begin{tabular}{ccccc}
    \toprule
            E1 &              E2 &              E3 &              E4 &              E5 \\
    \midrule
0.87$\pm$0.38 &   0.83$\pm$0.36 &   0.81$\pm$0.37 &   0.85$\pm$0.44 &   1.01$\pm$0.54 \\
0.28$\pm$0.40 &   0.26$\pm$0.40 &   0.26$\pm$0.40 &   0.27$\pm$0.39 &   0.32$\pm$0.40 \\
-0.15$\pm$0.44 &  -0.15$\pm$0.43 &  -0.15$\pm$0.43 &  -0.14$\pm$0.44 &  -0.09$\pm$0.43 \\
-0.43$\pm$0.48 &  -0.43$\pm$0.48 &  -0.43$\pm$0.48 &  -0.42$\pm$0.48 &  -0.36$\pm$0.47 \\
-0.47$\pm$0.42 &  -0.47$\pm$0.42 &  -0.46$\pm$0.43 &  -0.45$\pm$0.43 &  -0.34$\pm$0.59 \\
    \bottomrule
    \end{tabular}
    \end{subtable}

    \begin{subtable}[t]{.5\textwidth}
    \centering
    \caption{20-4-4-30-2.}
    \label{tab:sp3}
    \begin{tabular}{cccccc}
    \toprule
   &              E1 &              E2 &              E3 &              E4 &              E5 \\
    \midrule
L1 &   1.17$\pm$0.54 &   0.94$\pm$0.26 &   0.96$\pm$0.25 &   1.00$\pm$0.26 &  1.74$\pm$1.36 \\
L2 &   0.42$\pm$0.26 &   0.40$\pm$0.27 &   0.42$\pm$0.26 &   0.44$\pm$0.26 &  0.77$\pm$0.74 \\
L3 &   0.03$\pm$0.25 &   0.04$\pm$0.27 &   0.04$\pm$0.27 &   0.06$\pm$0.26 &  0.38$\pm$0.73 \\
L4 &  -0.23$\pm$0.26 &  -0.23$\pm$0.27 &  -0.22$\pm$0.27 &  -0.19$\pm$0.27 &  0.13$\pm$0.73 \\
L5 &  -0.06$\pm$0.62 &  -0.15$\pm$0.50 &  -0.25$\pm$0.33 &  -0.20$\pm$0.32 &  0.13$\pm$0.71 \\
    \bottomrule
    \end{tabular}
    \end{subtable}
    ~
    \begin{subtable}[t]{.5\textwidth}
    \centering
    \caption{25-8-8-40-2.}
    \label{tab:sp4}
    \begin{tabular}{ccccc}
    \toprule
            E1 &              E2 &              E3 &              E4 &               E5 \\
    \midrule
1.47$\pm$0.58 &   1.23$\pm$0.32 &   1.35$\pm$0.39 &  1.76$\pm$0.88 &  10.79$\pm$7.38 \\
0.71$\pm$0.32 &   0.63$\pm$0.31 &   0.66$\pm$0.31 &  1.07$\pm$0.78 &   5.82$\pm$4.27 \\
0.43$\pm$0.33 &   0.36$\pm$0.31 &   0.39$\pm$0.32 &  0.80$\pm$0.79 &   5.42$\pm$4.19 \\
0.05$\pm$0.34 &  -0.01$\pm$0.33 &   0.02$\pm$0.32 &  0.43$\pm$0.80 &   4.93$\pm$4.19 \\
-0.02$\pm$0.39 &  -0.04$\pm$0.43 &  -0.09$\pm$0.32 &  0.33$\pm$0.81 &   5.12$\pm$4.47 \\
    \bottomrule
    \end{tabular}
    \end{subtable}

    \begin{subtable}[t]{.5\textwidth}
    \centering
    \caption{30-4-4-50-2.}
    \label{tab:sp5}
    \begin{tabular}{cccccc}
    \toprule
   &              E1 &              E2 &              E3 &              E4 &              E5 \\
    \midrule
L1 &   2.41$\pm$2.26 &   1.69$\pm$0.96 &   1.60$\pm$0.86 &   2.04$\pm$1.44 &   5.33$\pm$5.37 \\
L2 &   0.64$\pm$0.38 &   0.53$\pm$0.25 &   0.55$\pm$0.24 &   0.72$\pm$0.35 &   1.49$\pm$1.40 \\
L3 &   0.06$\pm$0.24 &   0.04$\pm$0.23 &   0.06$\pm$0.24 &   0.20$\pm$0.26 &   0.53$\pm$0.81 \\
L4 &  -0.31$\pm$0.25 &  -0.35$\pm$0.22 &  -0.32$\pm$0.21 &  -0.21$\pm$0.26 &  -0.00$\pm$0.52 \\
L5 &  -0.40$\pm$0.31 &  -0.45$\pm$0.26 &  -0.44$\pm$0.24 &  -0.35$\pm$0.27 &   0.20$\pm$1.10 \\
    \bottomrule
    \end{tabular}
    \end{subtable}
    ~
    \begin{subtable}[t]{.5\textwidth}
    \centering
    \caption{30-6-6-50-1.5.}
    \label{tab:sp6}
    \begin{tabular}{ccccc}
    \toprule
                E1 &              E2 &              E3 &              E4 &              E5 \\
    \midrule
1.36$\pm$1.35 &   1.17$\pm$1.15 &   0.88$\pm$0.72 &   1.40$\pm$1.55 &  5.03$\pm$4.83 \\
0.18$\pm$0.44 &   0.08$\pm$0.19 &   0.04$\pm$0.15 &   0.08$\pm$0.14 &  1.82$\pm$2.09 \\
-0.37$\pm$0.16 &  -0.40$\pm$0.16 &  -0.38$\pm$0.16 &  -0.35$\pm$0.14 &  0.68$\pm$1.57 \\
-0.76$\pm$0.13 &  -0.76$\pm$0.15 &  -0.74$\pm$0.15 &  -0.70$\pm$0.15 &  0.24$\pm$1.50 \\
-0.92$\pm$0.13 &  -0.92$\pm$0.15 &  -0.89$\pm$0.13 &  -0.84$\pm$0.21 &  0.60$\pm$1.94 \\
    \bottomrule
    \end{tabular}
    \end{subtable}
\end{table*}

After the crossover and mutation operators, we repair the new individuals regarding the distribution of enemies. The crossover may generate levels that have more or fewer than the associated input parameter.
On the other hand, the mutation may transfer enemies to rooms that cannot have them, i.e., the goal room. If there are enemies in this room, we remove them. When the number of enemies is higher, we remove them, prioritizing the rooms with more of them.
If the number of enemies is lesser, we add them, prioritizing the non-empty rooms with fewer enemies.

Finally, the new individuals in the intermediate population are evaluated using an extended version of the fitness function in \cite{ref:pereira2021procedural}. Our function calculates three fitness factors.
First, we measure the distance of the input parameters and the generated level, i.e., how much closer is a generated level from the designer's input:
% New phrase
\begin{equation}
\begin{aligned}
f_{\text{goal}} =~& abs(G_{\mathit{rooms}} - L_{\mathit{rooms}})~+ \\
                  & abs(G_{\mathit{keys}} - L_{\mathit{keys}})~+ \\
                  & abs(G_{\mathit{locks}} - L_{\mathit{locks}})~+ \\
                  & abs(G_{\mathit{linear\_coefficient}} - L_{\mathit{linear\_coefficient}})~+ \\
                  & L_{\mathit{rooms}} - L_{\mathit{needed\_rooms}}~+ \\
                  & L_{\mathit{locks}} - L_{\mathit{needed\_locks}}
\end{aligned}
\end{equation}
where $G$ is the set of goals and $L$ is the set of the level's attributes (number of rooms, number of keys, number of locks, and linear coefficient); $needed\_rooms$ is calculated by an adaptation of a Depth-First Search algorithm, and; $needed\_locks$ calculated by an adaptation of an A$^{*}$ algorithm, both algorithms are described in  \cite{ref:pereira2021procedural}.

The second factor is an extension of the enemy sparsity equation introduced by Summerville et al. to evaluate the distribution of enemies in the 2D maps \cite{ref:summerville2017understanding}.
% New paragraph
This function encourages the dispersion of enemies in the levels' rooms (larger values mean more dispersion):
\begin{equation}
    f_{\text{es}} = \frac{\sum_{e \in E} (e_{x} - \mu_{x})^2 + (e_{y} - \mu_{y})^2}{\mathit{Number~of~Enemies}}
\end{equation}
where $e_{x}$ and $e_{y}$ are the x-position and y-position of an enemy $e$, $\mu_{x}$ and $\mu_{y}$ are the average x-position and y-position of all enemies, and $E$ is the set of enemies.
In the third term, we calculate the standard deviation of enemies in the rooms.
% New paragraph
This function encourages that the rooms have a balanced number of enemies (lower values imply enemies are evenly distributed):
\begin{equation}
    f_{\text{std}} = \sqrt{\frac{1}{N-2} \sum_{r \in R} (r_{\mathit{enemies}} - \mu_{\mathit{enemies}})^2}
\end{equation}
where $r$ is a room in the set of rooms $R$, $r_{enemies}$ is the number of enemies of a room, $\mu_{enemies}$ is the average number of enemies in the rooms, and $N$ is the number of rooms.
We subtract the starting and the goal rooms since they cannot have enemies. The final fitness expression follows:
\begin{equation}
    L_{\mathit{fitness}} = f_{\mathit{goal}} - f_{\mathit{es}} + f_{\mathit{std}}
\end{equation}
% $L_{{\f}itness}$ já é a função inteira, então tá redundante, agora se trocar '$L_{{\f}itness}$' por '$f_{goal}$' fica correto
We subtract the enemy sparsity $f_{es}$ because higher values are better for such metric, and we aim to minimize $f_{goal}$ and $f_{std}$ as well as our fitness function as a whole.

\section{Results} \label{sec:results} % 3pgs

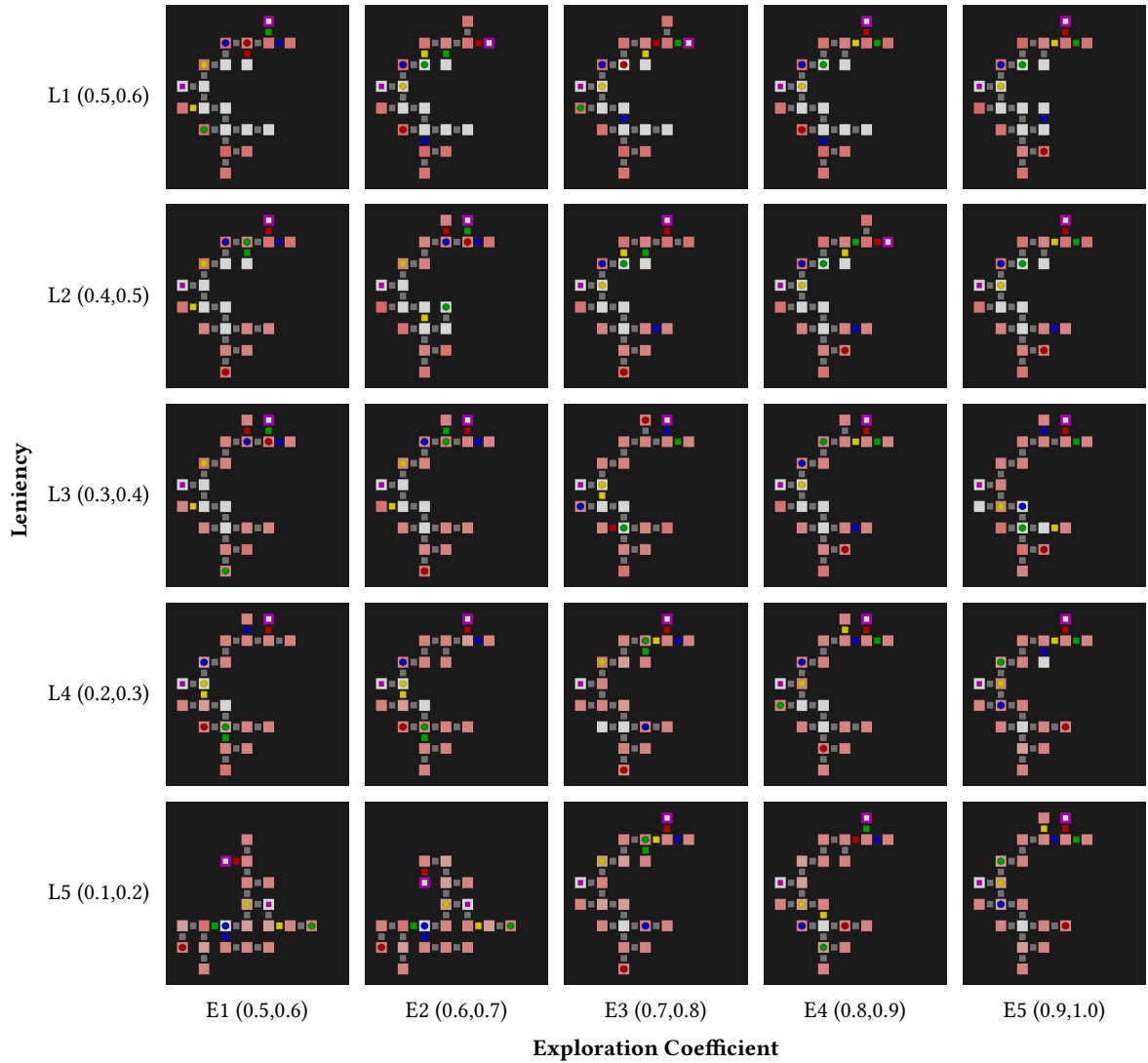
\begin{figure*}[ht]
    \centering
    
    \begin{tikzpicture}
    
    \node (l1e1) at (0,0) {
        \includegraphics[width=.14\textwidth]{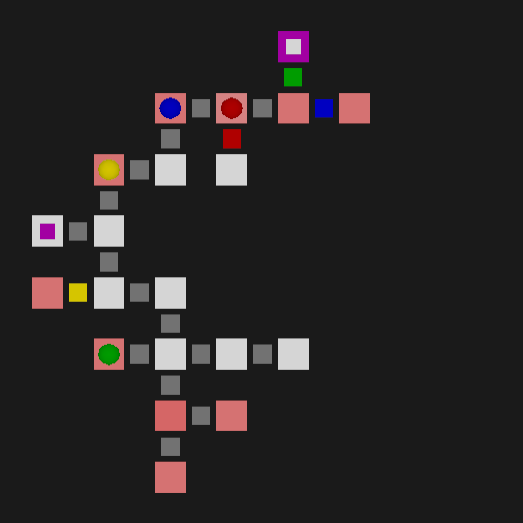}
    };
    \node (l1e2) [right = 0cm of l1e1] {
        \includegraphics[width=.14\textwidth]{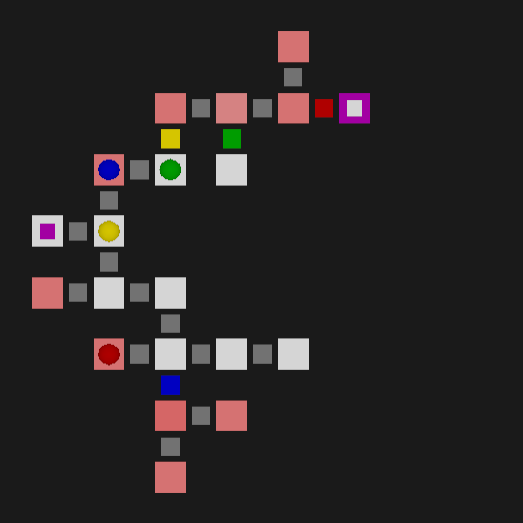}
    };
    \node (l1e3) [right = 0cm of l1e2] {
        \includegraphics[width=.14\textwidth]{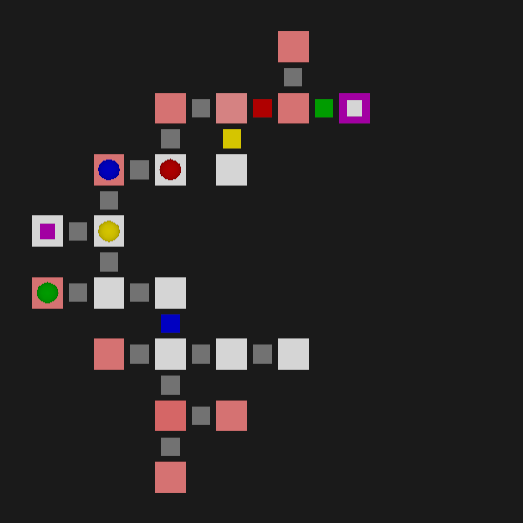}
    };
    \node (l1e4) [right = 0cm of l1e3] {
        \includegraphics[width=.14\textwidth]{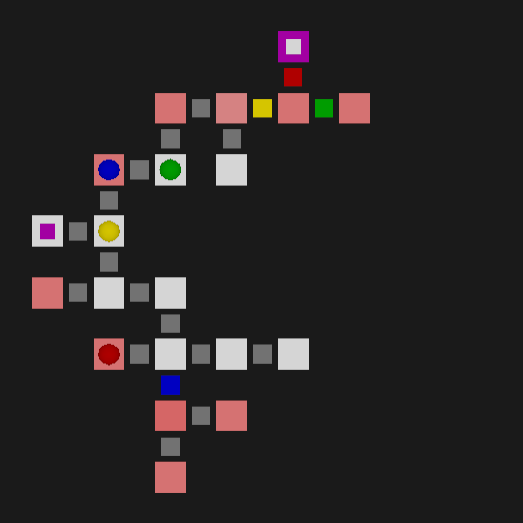}
    };
    \node (l1e5) [right = 0cm of l1e4] {
        \includegraphics[width=.14\textwidth]{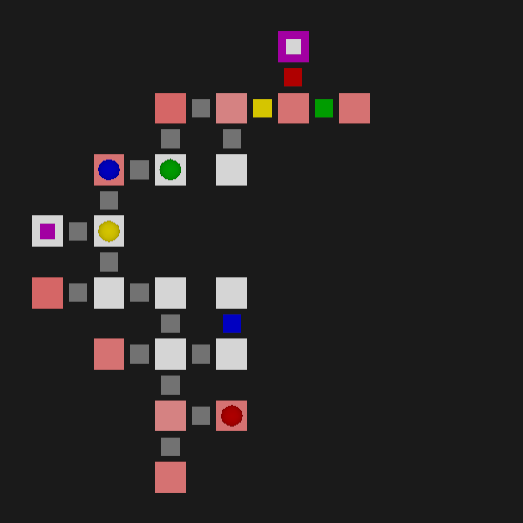}
    };
    \node (l2e1) [below = 0cm of l1e1] {
        \includegraphics[width=.14\textwidth]{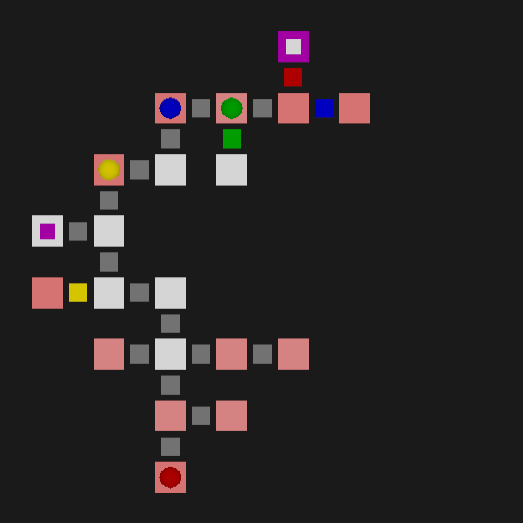}
    };
    \node (l2e2) [right = 0cm of l2e1] {
        \includegraphics[width=.14\textwidth]{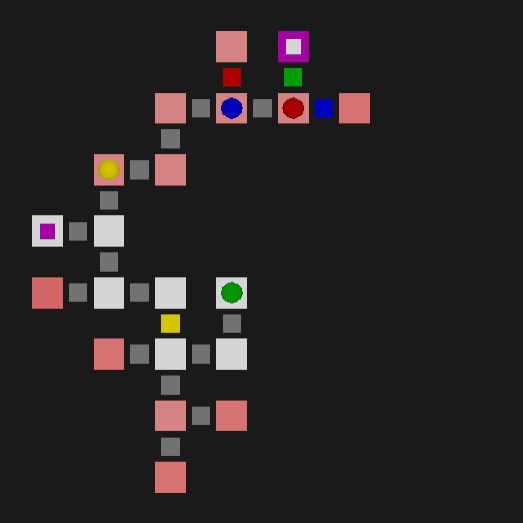}
    };
    \node (l2e3) [right = 0cm of l2e2] {
        \includegraphics[width=.14\textwidth]{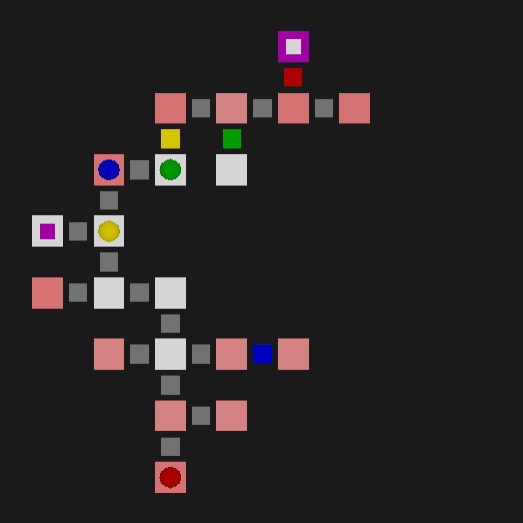}
    };
    \node (l2e4) [right = 0cm of l2e3] {
        \includegraphics[width=.14\textwidth]{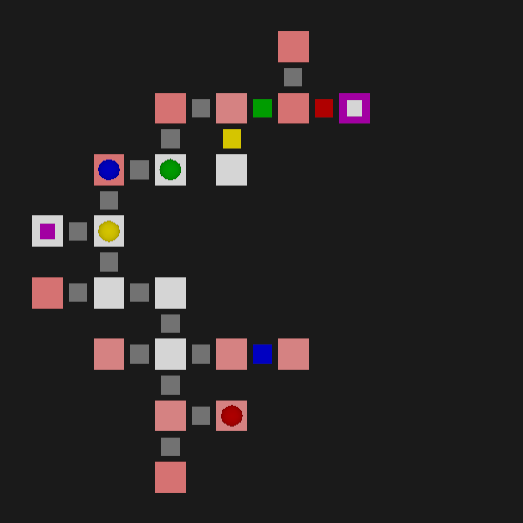}
    };
    \node (l2e5) [right = 0cm of l2e4] {
        \includegraphics[width=.14\textwidth]{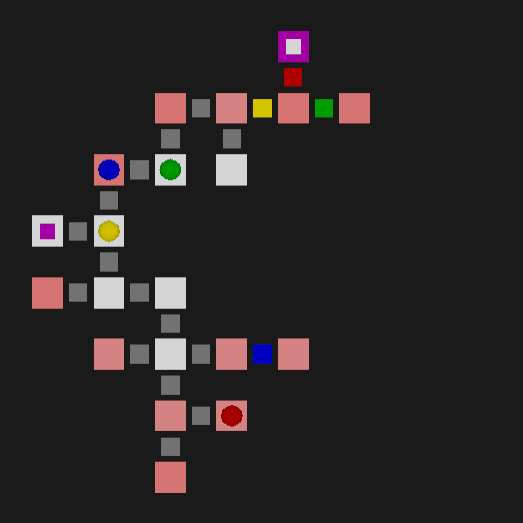}
    };
    \node (l3e1) [below = 0cm of l2e1] {
        \includegraphics[width=.14\textwidth]{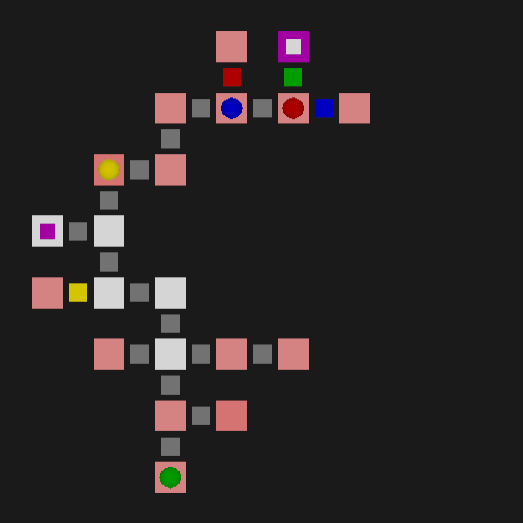}
    };
    \node (l3e2) [right = 0cm of l3e1] {
        \includegraphics[width=.14\textwidth]{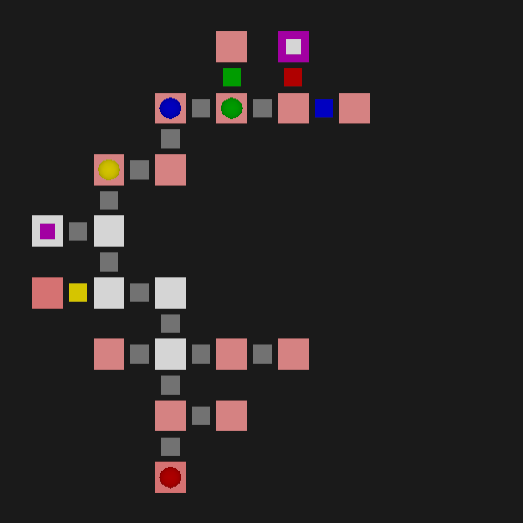}
    };
    \node (l3e3) [right = 0cm of l3e2] {
        \includegraphics[width=.14\textwidth]{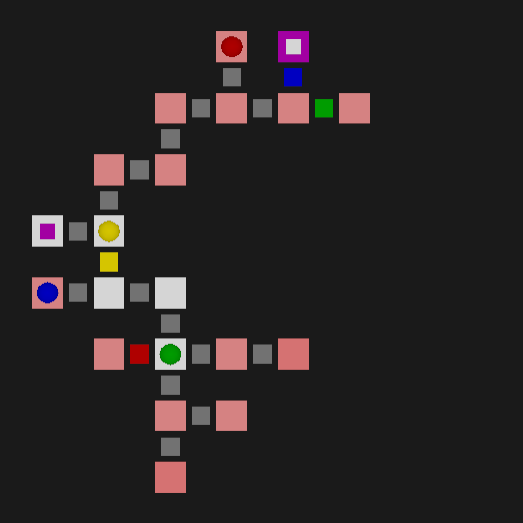}
    };
    \node (l3e4) [right = 0cm of l3e3] {
        \includegraphics[width=.14\textwidth]{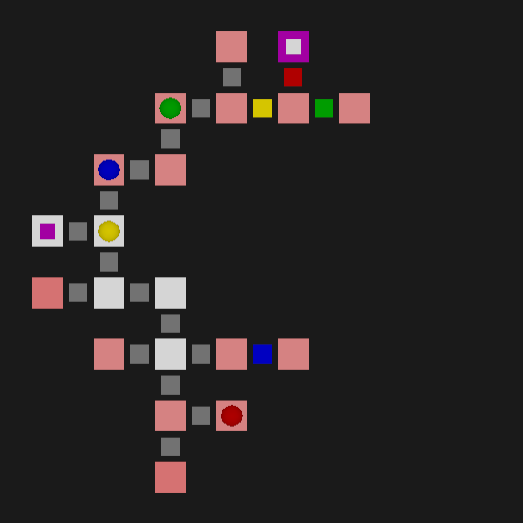}
    };
    \node (l3e5) [right = 0cm of l3e4] {
        \includegraphics[width=.14\textwidth]{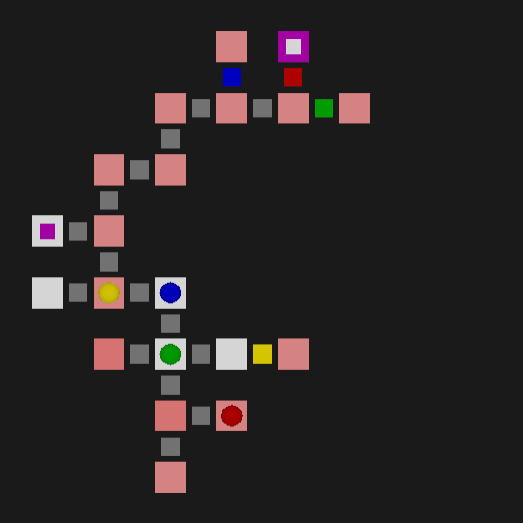}
    };
    \node (l4e1) [below = 0cm of l3e1] {
        \includegraphics[width=.14\textwidth]{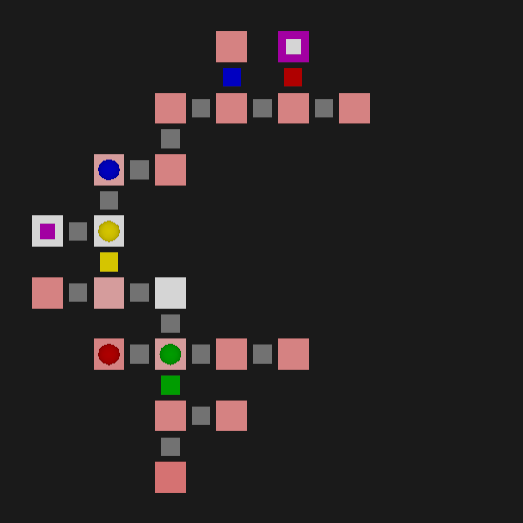}
    };
    \node (l4e2) [right = 0cm of l4e1] {
        \includegraphics[width=.14\textwidth]{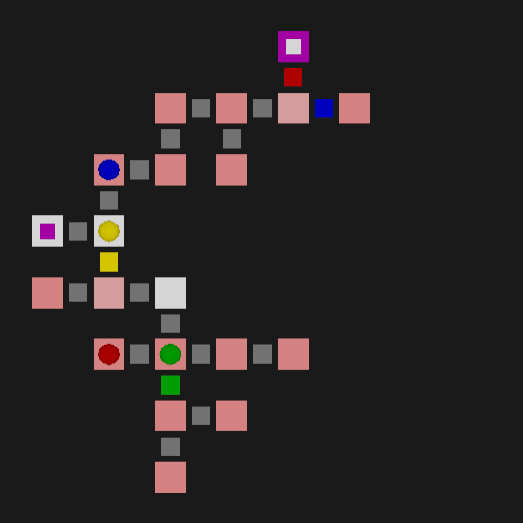}
    };
    \node (l4e3) [right = 0cm of l4e2] {
        \includegraphics[width=.14\textwidth]{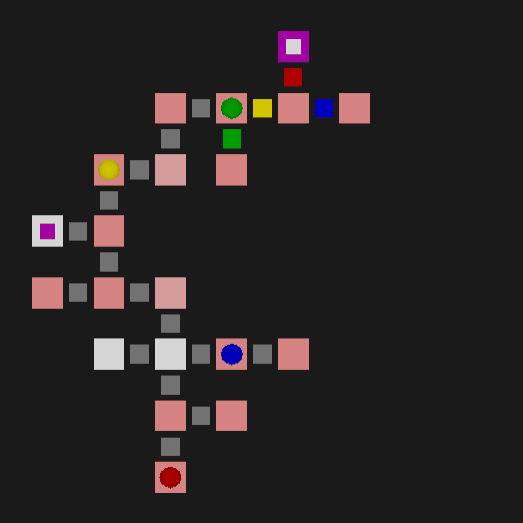}
    };
    \node (l4e4) [right = 0cm of l4e3] {
        \includegraphics[width=.14\textwidth]{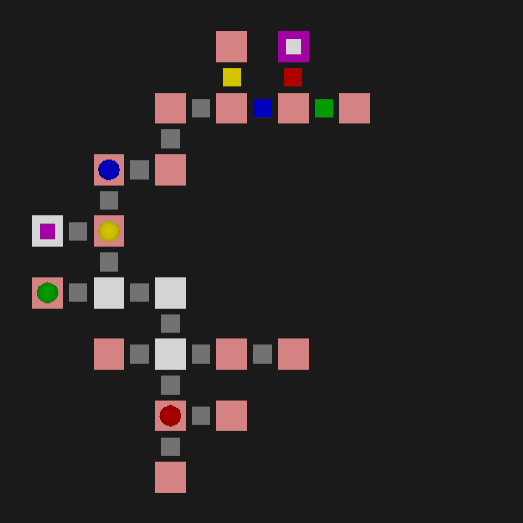}
    };
    \node (l4e5) [right = 0cm of l4e4] {
        \includegraphics[width=.14\textwidth]{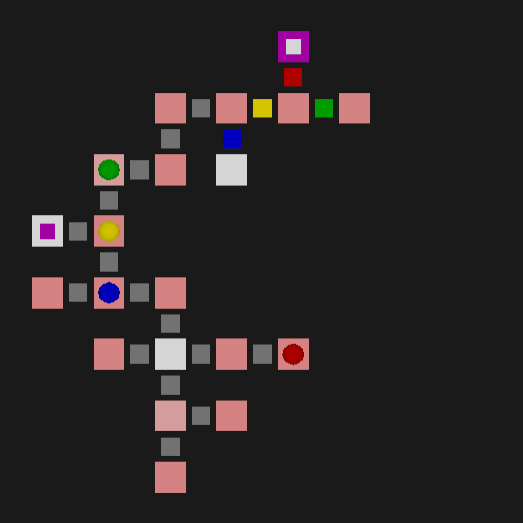}
    };
    \node (l5e1) [below = 0cm of l4e1] {
        \includegraphics[width=.14\textwidth]{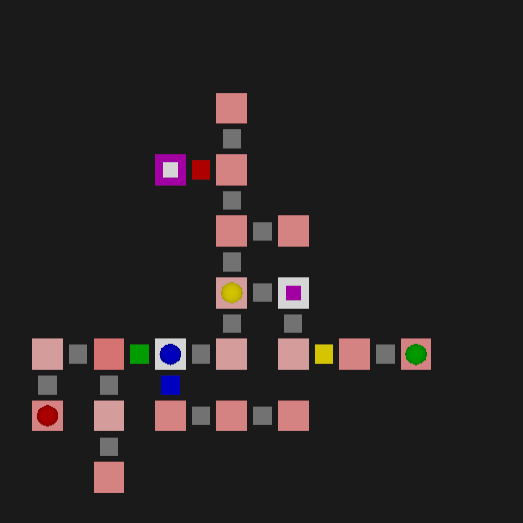}
    };
    \node (l5e2) [right = 0cm of l5e1] {
        \includegraphics[width=.14\textwidth]{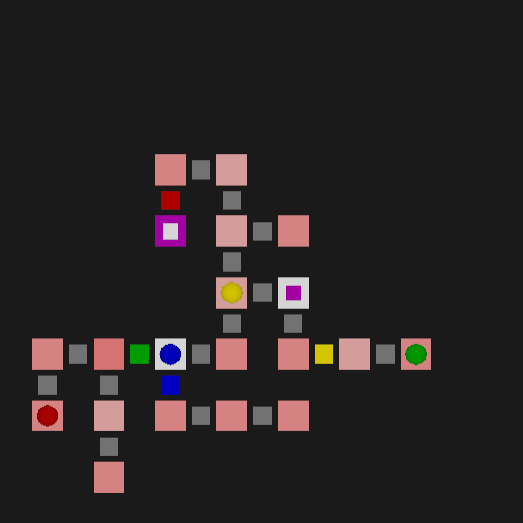}
    };
    \node (l5e3) [right = 0cm of l5e2] {
        \includegraphics[width=.14\textwidth]{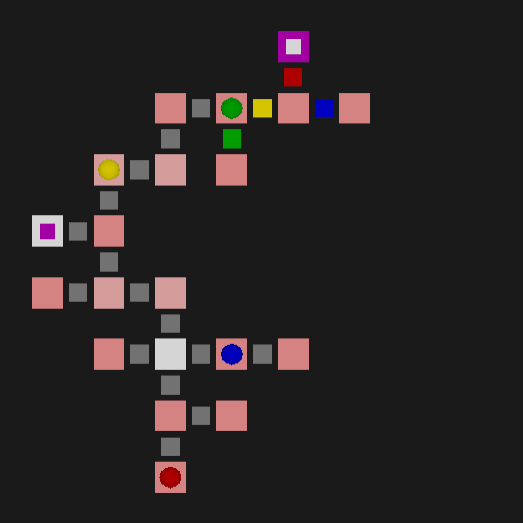}
    };
    \node (l5e4) [right = 0cm of l5e3] {
        \includegraphics[width=.14\textwidth]{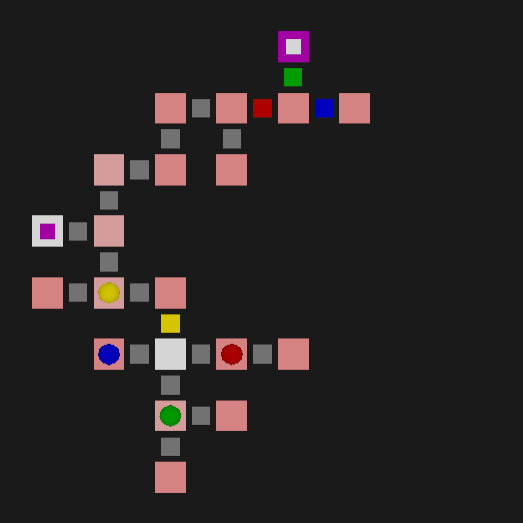}
    };
    \node (l5e5) [right = 0cm of l5e4] {
        \includegraphics[width=.14\textwidth]{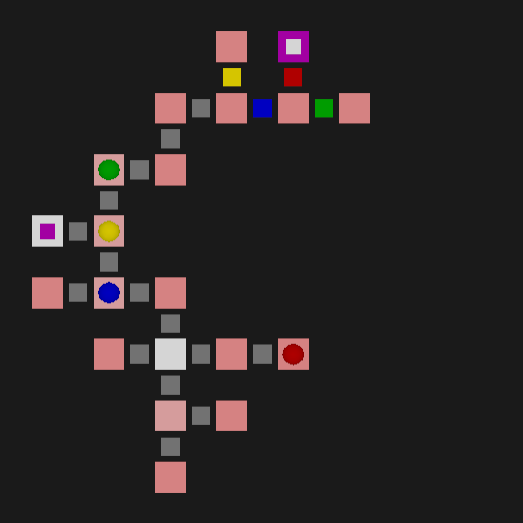}
    };
    
    % Axis
    \node (me1) [below=0cm of l5e1] {E1 (0.5,0.6)};
    \node (me2) [below=0cm of l5e2] {E2 (0.6,0.7)};
    \node (me3) [below=0cm of l5e3] {E3 (0.7,0.8)};
    \node (me4) [below=0cm of l5e4] {E4 (0.8,0.9)};
    \node (me5) [below=0cm of l5e5] {E5 (0.9,1.0)};
    \node (l1) [below=0cm of me3] {\textbf{Exploration Coefficient}};
    
    \node (ml1) [left=0cm of l1e1] {L1 (0.5,0.6)};
    \node (ml2) [left=0cm of l2e1] {L2 (0.4,0.5)};
    \node (ml3) [left=0cm of l3e1] {L3 (0.3,0.4)};
    \node (ml4) [left=0cm of l4e1] {L4 (0.2,0.3)};
    \node (ml5) [left=0cm of l5e1] {L5 (0.1,0.2)};
    \node (l2) [rotate=90, above left=0.58cm and 0cm of ml3] {\textbf{Leniency}};
    
    \end{tikzpicture}
    
    \caption{Example of a MAP-Elites population of levels with 20 rooms, 4 keys, 4 locks, 30 enemies, and linear coefficient equal to 2. Each table cell corresponds to an Elite. Descriptors for leniency values: L1 = (0.5,0.6), L2 = (0.4,0.5), L3 = (0.3,0.4), L4 = (0.2,0.3), L5 = (0.1,0.2). Descriptors for values of exploration coefficient: E1 = (0.5,0.6), E2 = (0.6,0.7), E3 = (0.7,0.8), E4 = (0.8,0.9), E5 = (0.9,1.0). The small squares represent corridors, and the bigger squares represent rooms. The white room with a purple square within it is the start room. The purple room with a white square within it is the goal room. White rooms have no enemies while red rooms have enemies within, the more intense the shade of red, the more enemies there are. Colored corridors are locked, and their keys are colored circles within rooms.}
    \label{fig:map-elite-result}
\end{figure*}

\begin{figure*}[ht]
    \centering
    
    \begin{subfigure}[t]{.23\textwidth}
    \centering
        \begin{tikzpicture}[every node/.style={scale=0.8}]
        \tikzstyle{node} = [minimum width=1.6cm, fill=gray!35!white]
        \begin{axis}
        [
            ylabel = Answers,
            xlabel = 5-point Likert Scale,
            ybar stacked,
            height=3.75cm,
            width=4.5cm,
            ymin = 0,
            ymax = 55,
            ytick={0,10,20,30,40,50},
            xtick={1,2,3,4,5},
            every axis plot/.append style={fill},
            nodes near coords={},
        ]
        \addplot [
            red!55!white!75!black,
            text=black,
            show sum on top,
        ] coordinates {
            (1,2)
            (2,18)
            (3,0)
            (4,0)
            (5,0)
        };
        \addplot [
            blue!55!white!75!black,
            text=black,
            show sum on top,
        ] coordinates {
            (1,0)
            (2,0)
            (3,21)
            (4,0)
            (5,0)
        };
        \addplot [
            green!55!white!75!black,
            text=black,
            show sum on top,
        ] coordinates {
            (1,0)
            (2,0)
            (3,0)
            (4,34)
            (5,46)
        };
        \end{axis}
        \node[node] at (0.7,2) {\textbf{AVG: 3.85}};
        \node[node] at (0.7,1.75) {\textbf{STD: 1.13}};
    \end{tikzpicture}
    \subcaption{Q1 - Was it fun?}
    \label{fig:q1}
    \end{subfigure}
    ~
    \begin{subfigure}[t]{.23\textwidth}
    \centering
    \begin{tikzpicture}[every node/.style={scale=0.8}]
        \tikzstyle{node} = [minimum width=1.6cm, fill=gray!35!white]
        \begin{axis}
        [
            ylabel = Answers,
            xlabel = 5-point Likert Scale,
            ybar stacked,
            height=3.75cm,
            width=4.5cm,
            ymin = 0,
            ymax = 55,
            ytick={0,10,20,30,40,50},
            xtick={1,2,3,4,5},
            every axis plot/.append style={fill},
            nodes near coords={},
        ]
        \addplot [
            red!55!white!75!black,
            text=black,
            show sum on top,
        ] coordinates {
            (1,22)
            (2,19)
            (3,0)
            (4,0)
            (5,0)
        };
        \addplot [
            blue!55!white!75!black,
            text=black,
            show sum on top,
        ] coordinates {
            (1,0)
            (2,0)
            (3,32)
            (4,0)
            (5,0)
        };
        \addplot [
            green!55!white!75!black,
            text=black,
            show sum on top,
        ] coordinates {
            (1,0)
            (2,0)
            (3,0)
            (4,26)
            (5,22)
        };
        \end{axis}
        \node[node] at (0.7,2) {\textbf{AVG: 3.05}};
        \node[node] at (0.7,1.75) {\textbf{STD: 1.35}};
    \end{tikzpicture}
    \subcaption{Q2 - Was it Difficult?}
    \label{fig:q2}
    \end{subfigure}
    ~
    \begin{subfigure}[t]{.23\textwidth}
    \centering
    \begin{tikzpicture}[every node/.style={scale=0.8}]
        \tikzstyle{node} = [minimum width=1.6cm, fill=gray!35!white]
        \begin{axis}
        [
            ylabel = Answers,
            xlabel = 5-point Likert Scale,
            ybar stacked,
            height=3.75cm,
            width=4.5cm,
            ymin = 0,
            ymax = 55,
            ytick={0,10,20,30,40,50},
            xtick={1,2,3,4,5},
            every axis plot/.append style={fill},
            nodes near coords={},
        ]
        \addplot [
            red!55!white!75!black,
            text=black,
            show sum on top,
        ] coordinates {
            (1,15)
            (2,22)
            (3,0)
            (4,0)
            (5,0)
        };
        \addplot [
            blue!55!white!75!black,
            text=black,
            show sum on top,
        ] coordinates {
            (1,0)
            (2,0)
            (3,23)
            (4,0)
            (5,0)
        };
        \addplot [
            green!55!white!75!black,
            text=black,
            show sum on top,
        ] coordinates {
            (1,0)
            (2,0)
            (3,0)
            (4,38)
            (5,23)
        };
        \end{axis}
        \node[node] at (0.7,2) {\textbf{AVG: 3.26}};
        \node[node] at (0.7,1.75) {\textbf{STD: 1.30}};
    \end{tikzpicture}
    \subcaption{Q3 - Challenge was right?}
    \label{fig:q3}
    \end{subfigure}
    ~
    \begin{subfigure}[t]{.23\textwidth}
    \centering
    \begin{tikzpicture}[every node/.style={scale=0.8}]
        \tikzstyle{node} = [minimum width=1.6cm, fill=gray!35!white]
        \begin{axis}
        [
            ylabel = Answers,
            xlabel = 5-point Likert Scale,
            ybar stacked,
            height=3.75cm,
            width=4.5cm,
            ymin = 0,
            ymax = 55,
            ytick={0,10,20,30,40,50},
            xtick={1,2,3,4,5},
            every axis plot/.append style={fill},
            nodes near coords={},
        ]
        \addplot [
            red!55!white!75!black,
            text=black,
            show sum on top,
        ] coordinates {
            (1,5)
            (2,11)
            (3,0)
            (4,0)
            (5,0)
        };
        \addplot [
            blue!55!white!75!black,
            text=black,
            show sum on top,
        ] coordinates {
            (1,0)
            (2,0)
            (3,23)
            (4,0)
            (5,0)
        };
        \addplot [
            green!55!white!75!black,
            text=black,
            show sum on top,
        ] coordinates {
            (1,0)
            (2,0)
            (3,0)
            (4,38)
            (5,44)
        };
        \end{axis}
        \node[node] at (0.7,2) {\textbf{AVG: 3.86}};
        \node[node] at (0.7,1.75) {\textbf{STD: 1.13}};
    \end{tikzpicture}
    \subcaption{Q4 - Liked the exploration?}
    \label{fig:q4}
    \end{subfigure}

    \begin{subfigure}[t]{.23\textwidth}
    \centering
    \begin{tikzpicture}[every node/.style={scale=0.8}]
        \tikzstyle{node} = [minimum width=1.6cm, fill=gray!35!white]
        \begin{axis}
        [
            ylabel = Answers,
            xlabel = 5-point Likert Scale,
            ybar stacked,
            height=3.75cm,
            width=4.5cm,
            ymin = 0,
            ymax = 55,
            ytick={0,10,20,30,40,50},
            xtick={1,2,3,4,5},
            every axis plot/.append style={fill},
            nodes near coords={},
        ]
        \addplot [
            red!55!white!75!black,
            text=black,
            show sum on top,
        ] coordinates {
            (1,7)
            (2,16)
            (3,0)
            (4,0)
            (5,0)
        };
        \addplot [
            blue!55!white!75!black,
            text=black,
            show sum on top,
        ] coordinates {
            (1,0)
            (2,0)
            (3,32)
            (4,0)
            (5,0)
        };
        \addplot [
            green!55!white!75!black,
            text=black,
            show sum on top,
        ] coordinates {
            (1,0)
            (2,0)
            (3,0)
            (4,28)
            (5,38)
        };
        \end{axis}
        \node[node] at (0.7,2) {\textbf{AVG: 3.61}};
        \node[node] at (0.7,1.75) {\textbf{STD: 1.22}};
    \end{tikzpicture}
    \subcaption{Q5 - Liked finding Keys?}
    \label{fig:q5}
    \end{subfigure}
    ~
    \begin{subfigure}[t]{.23\textwidth}
    \centering
    \begin{tikzpicture}[every node/.style={scale=0.8}]
        \tikzstyle{node} = [minimum width=1.6cm, fill=gray!35!white]
        \begin{axis}
        [
            ylabel = Answers,
            xlabel = 5-point Likert Scale,
            ybar stacked,
            height=3.75cm,
            width=4.5cm,
            ymin = 0,
            ymax = 55,
            ytick={0,10,20,30,40,50},
            xtick={1,2,3,4,5},
            every axis plot/.append style={fill},
            nodes near coords={},
        ]
        \addplot [
            red!55!white!75!black,
            text=black,
            show sum on top,
        ] coordinates {
            (1,31)
            (2,28)
            (3,0)
            (4,0)
            (5,0)
        };
        \addplot [
            blue!55!white!75!black,
            text=black,
            show sum on top,
        ] coordinates {
            (1,0)
            (2,0)
            (3,27)
            (4,0)
            (5,0)
        };
        \addplot [
            green!55!white!75!black,
            text=black,
            show sum on top,
        ] coordinates {
            (1,0)
            (2,0)
            (3,0)
            (4,13)
            (5,22)
        };
        \end{axis}
        \node[node] at (0.7,2) {\textbf{AVG: 2.72}};
        \node[node] at (0.7,1.75) {\textbf{STD: 1.42}};
    \end{tikzpicture}
    \subcaption{Q6 - Difficult to find exit?}
    \label{fig:q6}
    \end{subfigure}
    ~
    \begin{subfigure}[t]{.23\textwidth}
    \centering
    \begin{tikzpicture}[every node/.style={scale=0.8}]
        \tikzstyle{node} = [minimum width=1.6cm, fill=gray!35!white]
        \begin{axis}
        [
            ylabel = Answers,
            xlabel = 5-point Likert Scale,
            ybar stacked,
            height=3.75cm,
            width=4.5cm,
            ymin = 0,
            ymax = 55,
            ytick={0,10,20,30,40,50},
            xtick={1,2,3,4,5},
            every axis plot/.append style={fill},
            nodes near coords={},
        ]
        \addplot [
            red!55!white!75!black,
            text=black,
            show sum on top,
        ] coordinates {
            (1,17)
            (2,29)
            (3,0)
            (4,0)
            (5,0)
        };
        \addplot [
            blue!55!white!75!black,
            text=black,
            show sum on top,
        ] coordinates {
            (1,0)
            (2,0)
            (3,33)
            (4,0)
            (5,0)
        };
        \addplot [
            green!55!white!75!black,
            text=black,
            show sum on top,
        ] coordinates {
            (1,0)
            (2,0)
            (3,0)
            (4,18)
            (5,24)
        };
        \end{axis}
        \node[node] at (0.7,2) {\textbf{AVG: 3.02}};
        \node[node] at (0.7,1.75) {\textbf{STD: 1.32}};
    \end{tikzpicture}
    \subcaption{Q7 - Was created by humans?}
    \label{fig:q7}
    \end{subfigure}
    % ~
    % \begin{subfigure}[!t]{.23\textwidth}
    % \centering
    % \scriptsize
    % \vspace{-1.8cm}
    
    % % Tabela completa
    % % \resizebox{\textwidth}{!}{
    % % \begin{tabular}{cccccccc}
    % % \toprule
    % %     & \textbf{Q1} & \textbf{Q2} & \textbf{Q3} & \textbf{Q4} & \textbf{Q5} & \textbf{Q6} & \textbf{Q7} \\
    % % \midrule
    % % \textbf{AVG} & 3.85 & 3.05 & 3.26 & 3.86 & 3.61 & 2.72 & 3.02 \\
    % % \textbf{SD}  & 1.13 & 1.35 & 1.30 & 1.13 & 1.22 & 1.42 & 1.32 \\
    % % \bottomrule
    % % \end{tabular}}
    
    % \begin{tabular}{ccccc}
    % \toprule
    %     & \textbf{Q1} & \textbf{Q2} & \textbf{Q3} & \textbf{Q4} \\
    % \midrule
    % \textbf{AVG} & 3.85 & 3.05 & 3.26 & 3.86 \\
    % \textbf{SD}  & 1.13 & 1.35 & 1.30 & 1.13 \\
    % \bottomrule
    % \end{tabular}
    
    % \vspace{0.2cm}
    
    % \begin{tabular}{cccc}
    % \toprule
    %     & \textbf{Q5} & \textbf{Q6} & \textbf{Q7} \\
    % \midrule
    % \textbf{AVG} & 3.61 & 2.72 & 3.02 \\
    % \textbf{SD}  & 1.22 & 1.42 & 1.32 \\
    % \bottomrule
    % \end{tabular}
    
    % \vspace{0.1cm}
    
    % \subcaption{AVG and SD.}
    % \label{fig:avgsd}
    % \end{subfigure}
    
    \caption{Bar charts of answers of the 74 players for 121 levels. Each bar corresponds to the number of levels evaluated for the respective value of the five-point Likert scale. The questions from \autoref{subsec:feedback} were shortened for brevity.}
    \label{fig:feedback}
\end{figure*}
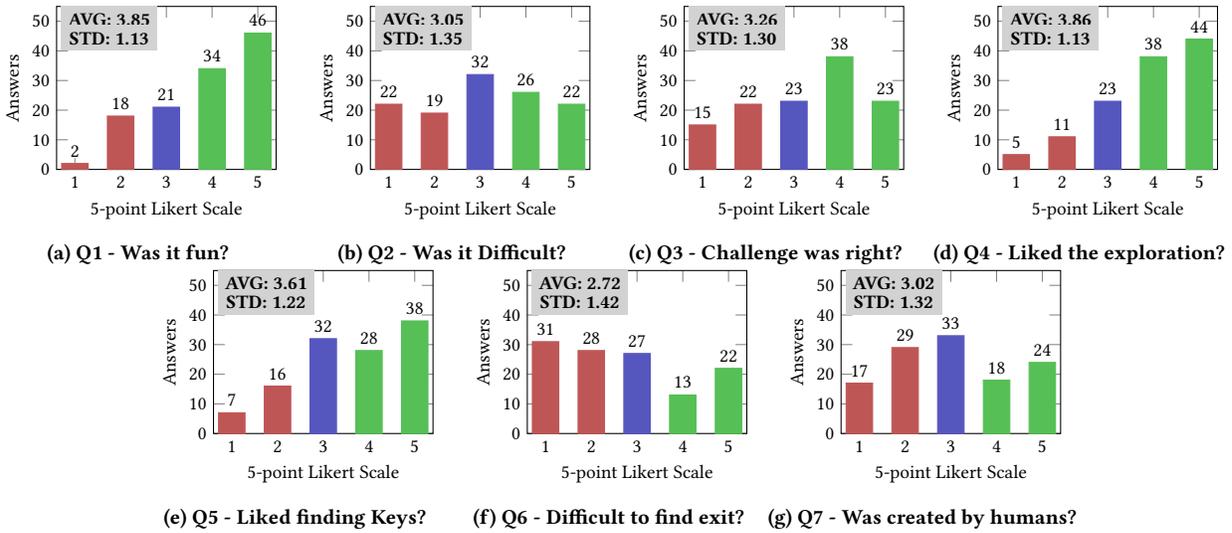

\begin{figure*}[ht]
    \centering
    
    \begin{subfigure}[t]{.23\textwidth}
    \centering
    \begin{tikzpicture}[every node/.style={scale=0.8}]
        \begin{axis}
        [
            ylabel = Answers,
            xlabel = 5-point Likert Scale,
            ybar stacked,
            height=3.2cm,
            width=4.5cm,
            ymin = 0,
            ymax = 25,
            ytick={0,5,10,15,20,25},
            xtick={1,2,3,4,5},
            every axis plot/.append style={fill},
            nodes near coords={},
        ]
        \addplot [
            red!55!white!75!black,
            text=black,
            show sum on top,
        ] coordinates {
            (1,3)
            (2,4)
            (3,0)
            (4,0)
            (5,0)
        };
        \addplot [
            blue!55!white!75!black,
            text=black,
            show sum on top,
        ] coordinates {
            (1,0)
            (2,0)
            (3,9)
            (4,0)
            (5,0)
        };
        \addplot [
            green!55!white!75!black,
            text=black,
            show sum on top,
        ] coordinates {
            (1,0)
            (2,0)
            (3,0)
            (4,19)
            (5,21)
        };
        \end{axis}
    \end{tikzpicture}
    \subcaption{E1.}
    \label{fig:fe_e1}
    \end{subfigure}
    ~
    \begin{subfigure}[t]{.23\textwidth}
    \centering
    \begin{tikzpicture}[every node/.style={scale=0.8}]
        \begin{axis}
        [
            ylabel = Answers,
            xlabel = 5-point Likert Scale,
            ybar stacked,
            height=3.2cm,
            width=4.5cm,
            ymin = 0,
            ymax = 25,
            ytick={0,5,10,15,20,25},
            xtick={1,2,3,4,5},
            every axis plot/.append style={fill},
            nodes near coords={},
        ]
        \addplot [
            red!55!white!75!black,
            text=black,
            show sum on top,
        ] coordinates {
            (1,0)
            (2,0)
            (3,0)
            (4,0)
            (5,0)
        };
        \addplot [
            blue!55!white!75!black,
            text=black,
            show sum on top,
        ] coordinates {
            (1,0)
            (2,0)
            (3,2)
            (4,0)
            (5,0)
        };
        \addplot [
            green!55!white!75!black,
            text=black,
            show sum on top,
        ] coordinates {
            (1,0)
            (2,0)
            (3,0)
            (4,1)
            (5,5)
        };
        \end{axis}
    \end{tikzpicture}
    \subcaption{E2.}
    \label{fig:fe_e2}
    \end{subfigure}
    ~
    \begin{subfigure}[t]{.23\textwidth}
    \centering
    \begin{tikzpicture}[every node/.style={scale=0.8}]
        \begin{axis}
        [
            ylabel = Answers,
            xlabel = 5-point Likert Scale,
            ybar stacked,
            height=3.2cm,
            width=4.5cm,
            ymin = 0,
            ymax = 25,
            ytick={0,5,10,15,20,25},
            xtick={1,2,3,4,5},
            every axis plot/.append style={fill},
            nodes near coords={},
        ]
        \addplot [
            red!55!white!75!black,
            text=black,
            show sum on top,
        ] coordinates {
            (1,1)
            (2,0)
            (3,0)
            (4,0)
            (5,0)
        };
        \addplot [
            blue!55!white!75!black,
            text=black,
            show sum on top,
        ] coordinates {
            (1,0)
            (2,0)
            (3,2)
            (4,0)
            (5,0)
        };
        \addplot [
            green!55!white!75!black,
            text=black,
            show sum on top,
        ] coordinates {
            (1,0)
            (2,0)
            (3,0)
            (4,5)
            (5,6)
        };
        \end{axis}
    \end{tikzpicture}
    \subcaption{E3.}
    \label{fig:fe_e3}
    \end{subfigure}
    ~
    \begin{subfigure}[t]{.23\textwidth}
    \centering
    \begin{tikzpicture}[every node/.style={scale=0.8}]
        \begin{axis}
        [
            ylabel = Answers,
            xlabel = 5-point Likert Scale,
            ybar stacked,
            height=3.2cm,
            width=4.5cm,
            ymin = 0,
            ymax = 25,
            ytick={0,5,10,15,20,25},
            xtick={1,2,3,4,5},
            every axis plot/.append style={fill},
            nodes near coords={},
        ]
        \addplot [
            red!55!white!75!black,
            text=black,
            show sum on top,
        ] coordinates {
            (1,0)
            (2,0)
            (3,0)
            (4,0)
            (5,0)
        };
        \addplot [
            blue!55!white!75!black,
            text=black,
            show sum on top,
        ] coordinates {
            (1,0)
            (2,0)
            (3,1)
            (4,0)
            (5,0)
        };
        \addplot [
            green!55!white!75!black,
            text=black,
            show sum on top,
        ] coordinates {
            (1,0)
            (2,0)
            (3,0)
            (4,2)
            (5,6)
        };
        \end{axis}
    \end{tikzpicture}
    \subcaption{E4.}
    \label{fig:fe_e4}
    \end{subfigure}
    
    \caption{Bar charts of answers for question Q5 (``I liked the challenge of finding the keys to this level'') of 57 players for 93 levels. These players answered, through a pre-questionnaire, they enjoy exploring. Each bar corresponds to the number of levels evaluated for the respective value of the five-point Likert scale. Each figure correspond to a descriptors for values of exploration coefficient: E1 = (0.5,0.6), E2 = (0.6,0.7), E3 = (0.7,0.8), E4 = (0.8,0.9), E5 = (0.9,1.0).}
    \label{fig:feedback_exp}
\end{figure*}
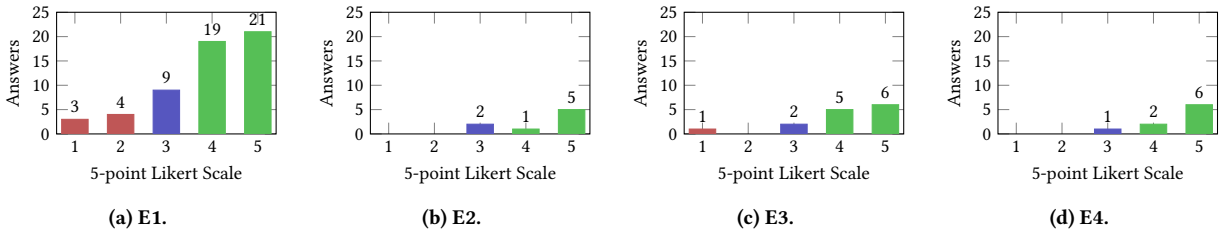

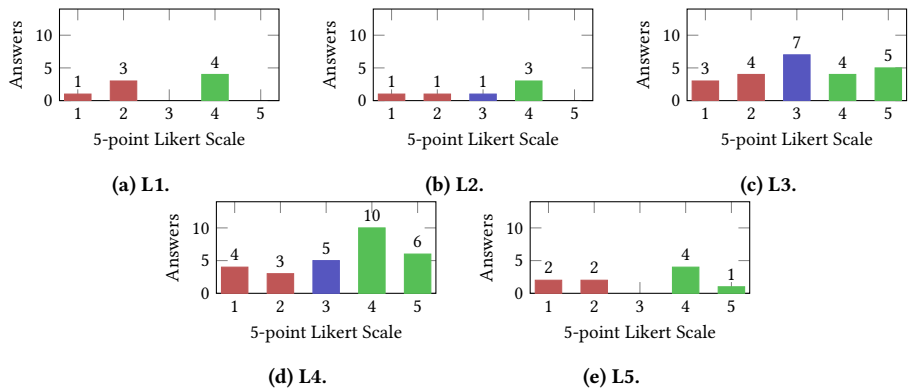
\begin{figure*}[ht]
    \centering
    
    \begin{subfigure}[t]{.23\textwidth}
    \centering
    \begin{tikzpicture}[every node/.style={scale=0.8}]
        \begin{axis}
        [
            ylabel = Answers,
            xlabel = 5-point Likert Scale,
            ybar stacked,
            height=2.8cm,
            width=4.5cm,
            ymin = 0,
            ymax = 14,
            ytick={0,5,10},
            xtick={1,2,3,4,5},
            every axis plot/.append style={fill},
            nodes near coords={},
        ]
        \addplot [
            red!55!white!75!black,
            text=black,
            show sum on top,
        ] coordinates {
            (1,1)
            (2,3)
            (3,0)
            (4,0)
            (5,0)
        };
        \addplot [
            blue!55!white!75!black,
            text=black,
            show sum on top,
        ] coordinates {
            (1,0)
            (2,0)
            (3,0)
            (4,0)
            (5,0)
        };
        \addplot [
            green!55!white!75!black,
            text=black,
            show sum on top,
        ] coordinates {
            (1,0)
            (2,0)
            (3,0)
            (4,4)
            (5,0)
        };
        \end{axis}
    \end{tikzpicture}
    \subcaption{L1.}
    \label{fig:fe_l1}
    \end{subfigure}
    ~
    \begin{subfigure}[t]{.23\textwidth}
    \centering
    \begin{tikzpicture}[every node/.style={scale=0.8}]
        \begin{axis}
        [
            ylabel = Answers,
            xlabel = 5-point Likert Scale,
            ybar stacked,
            height=2.8cm,
            width=4.5cm,
            ymin = 0,
            ymax = 14,
            ytick={0,5,10},
            xtick={1,2,3,4,5},
            every axis plot/.append style={fill},
            nodes near coords={},
        ]
        \addplot [
            red!55!white!75!black,
            text=black,
            show sum on top,
        ] coordinates {
            (1,1)
            (2,1)
            (3,0)
            (4,0)
            (5,0)
        };
        \addplot [
            blue!55!white!75!black,
            text=black,
            show sum on top,
        ] coordinates {
            (1,0)
            (2,0)
            (3,1)
            (4,0)
            (5,0)
        };
        \addplot [
            green!55!white!75!black,
            text=black,
            show sum on top,
        ] coordinates {
            (1,0)
            (2,0)
            (3,0)
            (4,3)
            (5,0)
        };
        \end{axis}
    \end{tikzpicture}
    \subcaption{L2.}
    \label{fig:fe_l2}
    \end{subfigure}
    ~
    \begin{subfigure}[t]{.23\textwidth}
    \centering
    \begin{tikzpicture}[every node/.style={scale=0.8}]
        \begin{axis}
        [
            ylabel = Answers,
            xlabel = 5-point Likert Scale,
            ybar stacked,
            height=2.8cm,
            width=4.5cm,
            ymin = 0,
            ymax = 14,
            ytick={0,5,10},
            xtick={1,2,3,4,5},
            every axis plot/.append style={fill},
            nodes near coords={},
        ]
        \addplot [
            red!55!white!75!black,
            text=black,
            show sum on top,
        ] coordinates {
            (1,3)
            (2,4)
            (3,0)
            (4,0)
            (5,0)
        };
        \addplot [
            blue!55!white!75!black,
            text=black,
            show sum on top,
        ] coordinates {
            (1,0)
            (2,0)
            (3,7)
            (4,0)
            (5,0)
        };
        \addplot [
            green!55!white!75!black,
            text=black,
            show sum on top,
        ] coordinates {
            (1,0)
            (2,0)
            (3,0)
            (4,4)
            (5,5)
        };
        \end{axis}
    \end{tikzpicture}
    \subcaption{L3.}
    \label{fig:fe_l3}
    \end{subfigure}
    
    \begin{subfigure}[t]{.23\textwidth}
    \centering
    \begin{tikzpicture}[every node/.style={scale=0.8}]
        \begin{axis}
        [
            ylabel = Answers,
            xlabel = 5-point Likert Scale,
            ybar stacked,
            height=2.8cm,
            width=4.5cm,
            ymin = 0,
            ymax = 14,
            ytick={0,5,10},
            xtick={1,2,3,4,5},
            every axis plot/.append style={fill},
            nodes near coords={},
        ]
        \addplot [
            red!55!white!75!black,
            text=black,
            show sum on top,
        ] coordinates {
            (1,4)
            (2,3)
            (3,0)
            (4,0)
            (5,0)
        };
        \addplot [
            blue!55!white!75!black,
            text=black,
            show sum on top,
        ] coordinates {
            (1,0)
            (2,0)
            (3,5)
            (4,0)
            (5,0)
        };
        \addplot [
            green!55!white!75!black,
            text=black,
            show sum on top,
        ] coordinates {
            (1,0)
            (2,0)
            (3,0)
            (4,10)
            (5,6)
        };
        \end{axis}
    \end{tikzpicture}
    \subcaption{L4.}
    \label{fig:fe_l4}
    \end{subfigure}
    ~
    \begin{subfigure}[t]{.23\textwidth}
    \centering
    \begin{tikzpicture}[every node/.style={scale=0.8}]
        \begin{axis}
        [
            ylabel = Answers,
            xlabel = 5-point Likert Scale,
            ybar stacked,
            height=2.8cm,
            width=4.5cm,
            ymin = 0,
            ymax = 14,
            ytick={0,5,10},
            xtick={1,2,3,4,5},
            every axis plot/.append style={fill},
            nodes near coords={},
        ]
        \addplot [
            red!55!white!75!black,
            text=black,
            show sum on top,
        ] coordinates {
            (1,2)
            (2,2)
            (3,0)
            (4,0)
            (5,0)
        };
        \addplot [
            blue!55!white!75!black,
            text=black,
            show sum on top,
        ] coordinates {
            (1,0)
            (2,0)
            (3,0)
            (4,0)
            (5,0)
        };
        \addplot [
            green!55!white!75!black,
            text=black,
            show sum on top,
        ] coordinates {
            (1,0)
            (2,0)
            (3,0)
            (4,4)
            (5,1)
        };
        \end{axis}
    \end{tikzpicture}
    \subcaption{L5.}
    \label{fig:fe_l5}
    \end{subfigure}
    
    \caption{Bar charts of answers for question Q3 (``The challenge was just right'') of 43 players for 74 levels. These players answered, through a pre-questionnaire, they enjoy battles. Each bar corresponds to the number of levels evaluated for the respective value of the five-point Likert scale. Each figure correspond to a descriptor for leniency values: L1 = (0.5,0.6), L2 = (0.4,0.5), L3 = (0.3,0.4), L4 = (0.2,0.3), L5 = (0.1,0.2).}
    \label{fig:feedback_len}
\end{figure*}

This section reports the computational results achieved by our approach, some level generated, and how human players evaluated our dungeons.

\subsection{Performance Results} % 1pg

We defined the evolutionary parameters empirically after comparing some range of values. The results comparing different configurations are available in a Google Sheets spreadsheet\footnote{Link to the spreadsheet: \href{https://docs.google.com/spreadsheets/d/1QmKPv8KyoavYy0jLWxQi07-itoTBv8VlJ8447X78Hvk}{https://docs.google.com/spreadsheets/d/1QmKPv8KyoavYy0\break jLWxQi07-itoTBv8VlJ8447X78Hvk}.}. After such evaluation, we set the following method's parameters: 25 individuals for initial population, 10\% for mutation rate, 100 individuals for intermediate population, 2-size for tournament selection, and 60 seconds as stop-criterion.

Next, we collected data from 30 executions of our method for six different sets of parameters to evaluate the algorithm performance. \autoref{tab:fitness_tables} shows the average and standard deviation of the fitness for each Elite (entry) of our MAP-Elites population. We observe that the fitness values tend to decrease as leniency decreases, which is expected because there are more safe rooms in L1 levels (50\% to 60\%) than L2 levels (40\% to 50\%), less in L3 levels, and so on. Moreover, L2 naturally presents their enemies distributed in more rooms than L1 levels; thus, increasing the enemy sparsity and decreasing the standard deviation of enemies.
% Eu escolhi 2 porque as elites L1-E1 e L1-E4 da tabela (e) não apresentaram um problema claro, a fitness ficou maior que 1 por causa de acúmulo de erros de fatores da f_goal
%The better levels present fitness values lower than two; therefore, we conclude that our algorithm converges for most Elites. 
By comparing tables \ref{tab:sp2} and \ref{tab:sp3}, we observe that increasing the linear coefficient decreases the dungeons' fitness.
That means that our algorithm works slightly better for lower linear coefficients.

E5 column in \autoref{tab:sp4} presents only subpar fitness values, and these results happen mainly due to the high number of locks at such a small level.
To be mapped in E5, the map coverage must fill from 90\% to 100\% of the level's rooms.
Nevertheless, the keys are usually closer to their locks; thus, they cannot be mapped.
We believe this result is caused mainly due to the crossover operation, which must ensure that both lock and key must be in the swapped branch.
Once lock and key are in the same branch, the coverage cannot fill 90\% of the level's rooms.
Thus, the levels found in the E5 column in \autoref{tab:sp4} have fewer locks than required by the input.

Finally, the Elite L1-E5 presents the worst fitness value in all the tables.
In this case, our algorithm should fill enemies in levels with 50\% up to 60\% safe rooms and ensure they present 90\% up to 100\% of exploration coefficient. Our algorithm struggled to find good results for such Elites.
Besides, this Elite has poor fitness values, particularly in the tables \ref{tab:sp5} and \ref{tab:sp6}. Such a result is an accumulation of bad values of the factors of $f_{goal}$, in which the main one is the number of rooms. Since our levels are randomly generated in the initial population, the difficulty of generating levels with a higher number of rooms is somewhat expected.

% \begin{figure}
%     \centering
%     \includegraphics[width=\linewidth, trim={3cm 0 3cm 0}]{images/results/GenerationError.png}
%     \caption{Average fitness and standard error of the best, worst and average of elites for each generation, averaged from 100 executions.}
%     \label{fig:fitnessEvolution}
% \end{figure}

\begin{figure}[!t]
    \centering
    \begin{tikzpicture}
        \begin{axis}[
            xlabel = Generations,
            ylabel = Fitness,
            height=5cm,
            width=7.5cm,
            scale only axis,
            xmin = -10,
            xmax = 620,
            ymin = -0.5,
            ymax = 7,
            ytick pos=left,
            ytick={0,1,2,3,4,5,6,7},
            xtick={0,100,200,300,400,500,600},
        ]
        \addplot[clip marker paths=true, color=blue!70!black, mark=none] table [x index=0, y index=1] {images/results/error/data.dat};
        \addplot[blue!70!black, error bars/.cd, y dir=plus, y explicit] table [x index=0, y index=1, y error index=2]{images/results/error/data.dat};
        \addplot[blue!70!black, error bars/.cd, y dir=minus, y explicit] table [x index=0, y index=1, y error index=2] {images/results/error/data.dat}; \label{plot:best}
        \addplot[clip marker paths=true, color=green!70!black, mark=none] table [x index=0, y index=3] {images/results/error/data.dat};
        \addplot[green!70!black, error bars/.cd, y dir=plus, y explicit] table [x index=0, y index=3, y error index=4]{images/results/error/data.dat};
        \addplot[green!70!black, error bars/.cd, y dir=minus, y explicit] table [x index=0, y index=3, y error index=4] {images/results/error/data.dat}; \label{plot:average}
        \addplot[clip marker paths=true, color=red!70!black, mark=none] table [x index=0, y index=5] {images/results/error/data.dat};
        \addplot[red!70!black, error bars/.cd, y dir=plus, y explicit] table [x index=0, y index=5, y error index=6]{images/results/error/data.dat};
        \addplot[red!70!black, error bars/.cd, y dir=minus, y explicit] table [x index=0, y index=5, y error index=6] {images/results/error/data.dat}; \label{plot:worst}
        
        \coordinate (legend) at (axis description cs:0.97,0.97);
        \end{axis}
        
        \matrix [
            draw,
            matrix of nodes,
            anchor=north east,
        ] at (legend) {
            \ref{plot:best} & Best Fitness \\
            \ref{plot:average} & Average Fitness \\
            \ref{plot:worst} & Worst Fitness \\
        };
    \end{tikzpicture}

    \caption{Average fitness and standard error of the best, worst and average of elites for each generation, averaged from 100 executions.}
    \label{fig:fitnessEvolution}
\end{figure}
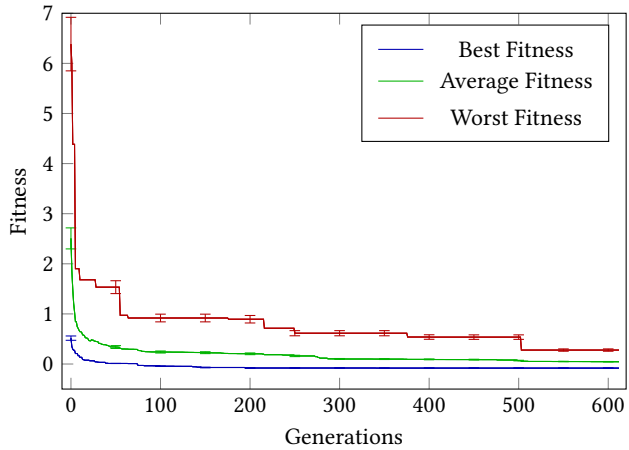

Finally, we test the consistency of the algorithm's convergence along with generations. \autoref{fig:fitnessEvolution} shows the evolution of the fitness and its standard error averaged from 100 executions using the inputs: 20 rooms, 4 keys, 4 locks, 30 enemies, and the linear coefficient equal to 2. Moreover, it shows the convergence's progression over 612 generations, which was the minimum number of generations that the 60s created over the 100 tests (60s being the parameter used for the other tests).

We can observe that the convergence is stable, with little standard error, especially after 500 generations. Our approach can converge to at least one good solution with its initial population (because of the preprocessing to guarantee some elites) and converge in less than 50 generations to an average of good solutions. Besides, with over 500 generations, even the worst elites are good.

\subsection{Generated Levels} % 1pg

\autoref{fig:map-elite-result} shows the result of an execution of our method. The figure shows the mapping performed by our algorithm, where we can see that the lower the leniency in the levels, the more the number of rooms with enemies (represented by squares with shades of red). The empty rooms (white squares) tend to be closer to each other, while rooms with enemies tend to be closer to the edges of the levels. We believe this behavior occurs due to the enemy sparsity, since it encourages the distribution of enemies regarding their position on the map. Besides,  there are stronger shades of red in the rooms with lesser leniency values, which is an expected result since we maintain the number of enemies in the whole level. The levels with high leniency degrees present more rooms with more enemies.

Regarding exploration coefficient, levels with lesser values for such metric present some keys closer to their locks; some are just in front of the lock they should open. Hence, as expected, the levels with higher exploration coefficients present a higher distance between the keys and their locks. For instance, the closer distance between a key and its lock in E5 levels is four rooms (e.g., in L1-E5 level). Nonetheless, there is at least a key in all the levels, far from its lock. In E1 levels, this key is the one that opens the goal room.

Moreover, considering only the positions of the rooms, the structure of most levels is very similar. Some similar levels differ only in terms of the enemies' position, such as L4-E3 and L5-E3. Locks and keys may also appear in the same positions (rooms), but they may change the required gameplay significantly. For instance, in level L2-E2, the player must collect the yellow key and open the yellow lock to collect the green key and open the goal room. In L2-E2, however, the player can access all the keys without unlocking any door. Nonetheless, this particular feature of chained locked-door missions is rare to appear in our approach; in \autoref{fig:map-elite-result} there are 7 out of 25 levels with this feature.

Although some of the levels are similar regarding room placement, we can observe that the algorithm worked as intended: a set of rooms different in both exploration and leniency were created, with most of them converging very close to the designer's needs and having interesting contents.

\subsection{Gameplay Feedback} \label{subsec:feedback}
% 1pg

Finally, we asked people to play a game prototype with the generated levels, and the players had to answer a questionnaire about each played level.
% Game Prototype
The game prototype is the same introduced by Pereira et al., but our locked-door missions are not generic, which means a key can open only a specific locked door  \cite{ref:pereira2021procedural}. Also, in our gameplay, the players must defeat enemies to progress, and our rooms may also have blocks that players can use to protect themselves from enemies. Thus, the gameplay in the game prototype, using levels procedurally generated by our MAP-Elites approach, advances from the original one in \cite{ref:pereira2021procedural}.

% Results
A total of 96 people played the levels, where 74 answered all the questions. They played 121 levels, randomly selected to feed the game prototype. After finishing a level, the players answered how much they agree or disagree, on a five-point Likert scale, with the following statements:

\vbox{
\begin{description}
% Q1 - 'PostQuestion 0' - O nível foi divertido de jogar
\item[Q1] The level was fun to play;
% Q2 - 'PostQuestion 1' - O nível foi difícil de completar
\item[Q2] The level was difficult to complete;
% Q3 - 'PostQuestion 3' - O desafio estava na medida certa
\item[Q3] The challenge was just right;
% Q4 - 'PostQuestion 5' - Eu gostei da quantidade de exploração disponível neste nível
\item[Q4] I liked the amount of exploration available on this level;
% Q5 - 'PostQuestion 6' - Eu gostei do desafio proporcionado em encontrar as chaves deste nível
\item[Q5] I liked the challenge of finding the keys to this level;
% Q6 - 'PostQuestion 7' - Foi difícil encontrar a saída deste nível
\item[Q6] It was difficult to find the exit/goal of this level;
% Q7 - 'PostQuestion 8' - Os níveis que joguei foram criados por humanos
\item[Q7] The levels I played were created by humans.
\end{description}
}

\autoref{fig:feedback} presents seven bar charts, each one with the answers to a question. Each chart summarizes their answers by presenting the average (AVG) and standard deviation (SD). The low SD ($\cong 1$) shows that the responses vary slightly. In \autoref{fig:q1}, the players had fun while playing 80 out of 121 levels, and only 20 did not enjoy it. \autoref{fig:q2} shows that most players did not have difficulty completing most of our levels (73 out of 121), and \autoref{fig:q3} shows the players felt that the challenge of 61 levels was just right, it was not good for 37 levels, and they felt neutral for 23 levels.

Players liked the exploration of 82 levels, as shown in \autoref{fig:q4}, they did not enjoy only 16 levels, and they were neutral for 23 levels.
We observe in \autoref{fig:q5} that the players liked the locked-doors puzzles in 66 levels; they did not like it in 23 levels and were neutral about it in 32 levels.
The players easily found the goal room in 59 levels in \autoref{fig:q6}; the goal room was difficult to find in only 35 levels.
Finally, \autoref{fig:q7} shows that the players believed that humans created 42 levels, 46 levels were generated by a PCG algorithm, and 33 had no sure.

Therefore, even while playing different levels, most players felt that playing the generated content was fun, with a balanced difficulty that brought a good challenge in combat against enemies and a good feeling of exploration in the dungeons. They liked the locked-doors puzzles while not finding it very difficult to find the exit.
Thus, our algorithm was able to bring quality and diversity to the solutions while also creating the content so that players could not accurately point out if an algorithm made it.

%The answers indicate satisfactory levels generated; however, they could improve, mainly regarding the challenge. 
%Besides, most of our levels look not created carefully as humans usually do.

% Extra

Besides the questionnaire to evaluate levels, we also asked the players if they enjoyed exploring and battling during their gameplay.
\autoref{fig:feedback_exp} shows the feedback of the exploration of levels that were played by players that enjoy exploring by class of exploration coefficient. Most players played E1 levels, and no one played E5 levels; however, most enjoyed playing all the levels independent of the value of the exploration coefficient.
\autoref{fig:feedback_len} shows the feedback of levels of the players that enjoy battling. The results vary more in these charts, with most players playing L4 levels and agreeing with the challenge in the levels they played. Regarding the levels with the remaining leniency values, we cannot declare if they presented the just right amount of challenge, since the number of players who agreed and disagreed with that is too close.

\section{Conclusion} \label{sec:conclusion} % 0.5gps

% CLAUDIO: pelo que eu entendi, no site do FDG diz que pode ter uma nona página desde que sejam só referências: https://FDG-2022.sigevo.org/Call-for-Papers#Paper_Categories

This paper introduced an illumination approach that extended the work presented by Pereira et al. \cite{ref:pereira2021procedural}.
Our contributions advance the method by orchestrating enemies within the levels (level facet) and the locked-door missions (narrative facet) through their illumination with the MAP-Elites approach.
The experiments show that our evolutionary level generation approach is stable, concerning the standard deviation of the fitness, and converges all the dungeons on the map in many executions.
Regarding the experiments with people, the results show that most players positively answered the levels we generated.
The players enjoyed the levels created by our algorithms and could not indicate if an algorithm created the levels.
Thus, our approach maintains the dungeon quality -- once our results corroborate the results of the original method \cite{ref:pereira2021procedural} -- and goes beyond by providing a set of diverse levels.
As future works, we intend to add a novelty score to allow more distinct levels in the map in terms of level structure as proposed in \cite{ref:conti2017improving}.

% % Comentar antes de submeter
% \begin{acks}
% We acknowledge the financial support of the National Council for Scientific and Technological Development (CNPq).
% \end{acks}

\bibliographystyle{ACM-Reference-Format}
\bibliography{main}


\begin{thebibliography}{25}

%%% the Web address.

\ifx \showCODEN    \undefined \def \showCODEN     #1{\unskip}     \fi
\ifx \showDOI      \undefined \def \showDOI       #1{#1}\fi
\ifx \showISBNx    \undefined \def \showISBNx     #1{\unskip}     \fi
\ifx \showISBNxiii \undefined \def \showISBNxiii  #1{\unskip}     \fi
\ifx \showISSN     \undefined \def \showISSN      #1{\unskip}     \fi
\ifx \showLCCN     \undefined \def \showLCCN      #1{\unskip}     \fi
\ifx \shownote     \undefined \def \shownote      #1{#1}          \fi
\ifx \showarticletitle \undefined \def \showarticletitle #1{#1}   \fi
\ifx \showURL      \undefined \def \showURL       {\relax}        \fi
% invisible to TeX
\providecommand\bibfield[2]{#2}
\providecommand\bibinfo[2]{#2}
\providecommand\natexlab[1]{#1}
\providecommand\showeprint[2][]{arXiv:#2}

\bibitem[\protect\citeauthoryear{Alvarez, Dahlskog, Font, and Togelius}{Alvarez
  et~al\mbox{.}}{2019}]%
        {ref:alvarez2019empowering}
\bibfield{author}{\bibinfo{person}{Alberto Alvarez}, \bibinfo{person}{Steve
  Dahlskog}, \bibinfo{person}{Jose Font}, {and} \bibinfo{person}{Julian
  Togelius}.} \bibinfo{year}{2019}\natexlab{}.
\newblock \showarticletitle{Empowering quality diversity in dungeon design with
  interactive constrained MAP-Elites}. In \bibinfo{booktitle}{\emph{2019 IEEE
  Conference on Games (CoG)}}. IEEE, \bibinfo{pages}{1--8}.
\newblock


\bibitem[\protect\citeauthoryear{Baldwin, Dahlskog, Font, and Holmberg}{Baldwin
  et~al\mbox{.}}{2017}]%
        {ref:baldwin2017mixed}
\bibfield{author}{\bibinfo{person}{Alexander Baldwin}, \bibinfo{person}{Steve
  Dahlskog}, \bibinfo{person}{Jose~M Font}, {and} \bibinfo{person}{Johan
  Holmberg}.} \bibinfo{year}{2017}\natexlab{}.
\newblock \showarticletitle{Mixed-initiative procedural generation of dungeons
  using game design patterns}. In \bibinfo{booktitle}{\emph{Computational
  Intelligence and Games (CIG), 2017 IEEE Conference on}}. IEEE,
  \bibinfo{pages}{25--32}.
\newblock


\bibitem[\protect\citeauthoryear{Charity, Green, Khalifa, and Togelius}{Charity
  et~al\mbox{.}}{2020}]%
        {ref:charity2020mech}
\bibfield{author}{\bibinfo{person}{Megan Charity},
  \bibinfo{person}{Michael~Cerny Green}, \bibinfo{person}{Ahmed Khalifa}, {and}
  \bibinfo{person}{Julian Togelius}.} \bibinfo{year}{2020}\natexlab{}.
\newblock \showarticletitle{Mech-Elites: Illuminating the Mechanic Space of
  GVGAI}.
\newblock \bibinfo{journal}{\emph{arXiv preprint arXiv:2002.04733}}
  (\bibinfo{year}{2020}).
\newblock


\bibitem[\protect\citeauthoryear{Conti, Madhavan, Such, Lehman, Stanley, and
  Clune}{Conti et~al\mbox{.}}{2017}]%
        {ref:conti2017improving}
\bibfield{author}{\bibinfo{person}{Edoardo Conti}, \bibinfo{person}{Vashisht
  Madhavan}, \bibinfo{person}{Felipe~Petroski Such}, \bibinfo{person}{Joel
  Lehman}, \bibinfo{person}{Kenneth~O Stanley}, {and} \bibinfo{person}{Jeff
  Clune}.} \bibinfo{year}{2017}\natexlab{}.
\newblock \showarticletitle{Improving exploration in evolution strategies for
  deep reinforcement learning via a population of novelty-seeking agents}.
\newblock \bibinfo{journal}{\emph{arXiv preprint arXiv:1712.06560}}
  (\bibinfo{year}{2017}).
\newblock


\bibitem[\protect\citeauthoryear{De~Kegel and Haahr}{De~Kegel and
  Haahr}{2019}]%
        {ref:de2019procedural}
\bibfield{author}{\bibinfo{person}{Barbara De~Kegel} {and}
  \bibinfo{person}{Mads Haahr}.} \bibinfo{year}{2019}\natexlab{}.
\newblock \showarticletitle{Procedural puzzle generation: a survey}.
\newblock \bibinfo{journal}{\emph{IEEE Transactions on Games}}
  \bibinfo{volume}{12}, \bibinfo{number}{1} (\bibinfo{year}{2019}),
  \bibinfo{pages}{21--40}.
\newblock


\bibitem[\protect\citeauthoryear{{Digital Sun}}{{Digital Sun}}{2018}]%
        {moonlighter}
\bibfield{author}{\bibinfo{person}{{Digital Sun}}.}
  \bibinfo{year}{2018}\natexlab{}.
\newblock \bibinfo{title}{Moonlighter}.
\newblock
\newblock
\urldef\tempurl%
\url{http://moonlighterthegame.com/.}
\showURL{%
\tempurl}
\newblock
\shownote{Accessed in: 2020-07-25}.


\bibitem[\protect\citeauthoryear{Dormans}{Dormans}{2010}]%
        {ref:dormans2010adventures}
\bibfield{author}{\bibinfo{person}{Joris Dormans}.}
  \bibinfo{year}{2010}\natexlab{}.
\newblock \showarticletitle{Adventures in level design: generating missions and
  spaces for action adventure games}. In \bibinfo{booktitle}{\emph{Proceedings
  of the 2010 workshop on procedural content generation in games}}. ACM,
  \bibinfo{pages}{1}.
\newblock


\bibitem[\protect\citeauthoryear{Dormans}{Dormans}{2011}]%
        {ref:dormans2011level}
\bibfield{author}{\bibinfo{person}{Joris Dormans}.}
  \bibinfo{year}{2011}\natexlab{}.
\newblock \showarticletitle{Level design as model transformation: a strategy
  for automated content generation}. In \bibinfo{booktitle}{\emph{Proceedings
  of the 2nd International Workshop on Procedural Content Generation in
  Games}}. \bibinfo{pages}{1--8}.
\newblock


\bibitem[\protect\citeauthoryear{Dormans and Bakkes}{Dormans and
  Bakkes}{2011}]%
        {ref:dormans2011generating}
\bibfield{author}{\bibinfo{person}{Joris Dormans} {and} \bibinfo{person}{Sander
  Bakkes}.} \bibinfo{year}{2011}\natexlab{}.
\newblock \showarticletitle{Generating missions and spaces for adaptable play
  experiences}.
\newblock \bibinfo{journal}{\emph{IEEE Transactions on Computational
  Intelligence and AI in Games}} \bibinfo{volume}{3}, \bibinfo{number}{3}
  (\bibinfo{year}{2011}), \bibinfo{pages}{216--228}.
\newblock


\bibitem[\protect\citeauthoryear{Gellel and Sweetser}{Gellel and
  Sweetser}{2020}]%
        {ref:gellel2020hybrid}
\bibfield{author}{\bibinfo{person}{Alexander Gellel} {and}
  \bibinfo{person}{Penny Sweetser}.} \bibinfo{year}{2020}\natexlab{}.
\newblock \showarticletitle{A Hybrid Approach to Procedural Generation of
  Roguelike Video Game Levels}. In \bibinfo{booktitle}{\emph{International
  Conference on the Foundations of Digital Games}}. \bibinfo{pages}{1--10}.
\newblock


\bibitem[\protect\citeauthoryear{Gravina, Khalifa, Liapis, Togelius, and
  Yannakakis}{Gravina et~al\mbox{.}}{2019}]%
        {ref:gravina2019procedural}
\bibfield{author}{\bibinfo{person}{Daniele Gravina}, \bibinfo{person}{Ahmed
  Khalifa}, \bibinfo{person}{Antonios Liapis}, \bibinfo{person}{Julian
  Togelius}, {and} \bibinfo{person}{Georgios~N Yannakakis}.}
  \bibinfo{year}{2019}\natexlab{}.
\newblock \showarticletitle{Procedural content generation through quality
  diversity}. In \bibinfo{booktitle}{\emph{2019 IEEE Conference on Games
  (CoG)}}. IEEE, \bibinfo{pages}{1--8}.
\newblock


\bibitem[\protect\citeauthoryear{{Hello Games}}{{Hello Games}}{2018}]%
        {ref:nomanssky}
\bibfield{author}{\bibinfo{person}{{Hello Games}}.}
  \bibinfo{year}{2018}\natexlab{}.
\newblock \bibinfo{title}{No Man's Sky}.
\newblock
\newblock
\urldef\tempurl%
\url{https://www.nomanssky.com.}
\showURL{%
\tempurl}
\newblock
\shownote{Accessed in: 2021-11-09}.


\bibitem[\protect\citeauthoryear{Karavolos, Liapis, and Yannakakis}{Karavolos
  et~al\mbox{.}}{2018}]%
        {karavolos2018facets}
\bibfield{author}{\bibinfo{person}{Daniel Karavolos}, \bibinfo{person}{Antonios
  Liapis}, {and} \bibinfo{person}{Georgios Yannakakis}.}
  \bibinfo{year}{2018}\natexlab{}.
\newblock \showarticletitle{A Multi-Faceted Surrogate Model for Search-based
  Procedural Content Generation}.
\newblock \bibinfo{journal}{\emph{IEEE TRANSACTIONS ON GAMES}}
  (\bibinfo{year}{2018}).
\newblock
Issue X.
\urldef\tempurl%
\url{https://wiki.teamfortress.com/wiki/Classes}
\showURL{%
\tempurl}


\bibitem[\protect\citeauthoryear{Liapis}{Liapis}{2017}]%
        {ref:liapis2017multi}
\bibfield{author}{\bibinfo{person}{Antonios Liapis}.}
  \bibinfo{year}{2017}\natexlab{}.
\newblock \showarticletitle{Multi-segment evolution of dungeon game levels}. In
  \bibinfo{booktitle}{\emph{Proceedings of the Genetic and Evolutionary
  Computation Conference}}. ACM, \bibinfo{pages}{203--210}.
\newblock


\bibitem[\protect\citeauthoryear{Liapis, Yannakakis, and Togelius}{Liapis
  et~al\mbox{.}}{2013}]%
        {ref:liapis2013towards}
\bibfield{author}{\bibinfo{person}{Antonios Liapis}, \bibinfo{person}{Georgios
  Yannakakis}, {and} \bibinfo{person}{Julian Togelius}.}
  \bibinfo{year}{2013}\natexlab{}.
\newblock \showarticletitle{Towards a generic method of evaluating game
  levels}. In \bibinfo{booktitle}{\emph{Proceedings of the AAAI Conference on
  Artificial Intelligence and Interactive Digital Entertainment}},
  Vol.~\bibinfo{volume}{9}.
\newblock


\bibitem[\protect\citeauthoryear{Liapis, Yannakakis, Nelson, Preuss, and
  Bidarra}{Liapis et~al\mbox{.}}{2019}]%
        {liapis2019maestro}
\bibfield{author}{\bibinfo{person}{Antonios Liapis},
  \bibinfo{person}{Georgios~N. Yannakakis}, \bibinfo{person}{Mark~J. Nelson},
  \bibinfo{person}{Mike Preuss}, {and} \bibinfo{person}{Rafael Bidarra}.}
  \bibinfo{year}{2019}\natexlab{}.
\newblock \showarticletitle{Orchestrating game generation}.
\newblock \bibinfo{journal}{\emph{IEEE Transactions on Games}}
  \bibinfo{volume}{11} (\bibinfo{date}{3} \bibinfo{year}{2019}),
  \bibinfo{pages}{48--68}.
\newblock
Issue 1.
\showISSN{24751510}
\urldef\tempurl%
\url{https://doi.org/10.1109/TG.2018.2870876}
\showDOI{\tempurl}


\bibitem[\protect\citeauthoryear{Mouret and Clune}{Mouret and Clune}{2015}]%
        {ref:mouret2015illuminating}
\bibfield{author}{\bibinfo{person}{Jean-Baptiste Mouret} {and}
  \bibinfo{person}{Jeff Clune}.} \bibinfo{year}{2015}\natexlab{}.
\newblock \showarticletitle{Illuminating search spaces by mapping elites}.
\newblock \bibinfo{journal}{\emph{arXiv preprint arXiv:1504.04909}}
  (\bibinfo{year}{2015}).
\newblock


\bibitem[\protect\citeauthoryear{Pereira, de~Souza~Prado, Lopes, and
  Toledo}{Pereira et~al\mbox{.}}{2021}]%
        {ref:pereira2021procedural}
\bibfield{author}{\bibinfo{person}{Leonardo~Tortoro Pereira},
  \bibinfo{person}{Paulo~Victor de Souza~Prado},
  \bibinfo{person}{Rafael~Miranda Lopes}, {and} \bibinfo{person}{Claudio
  Fabiano~Motta Toledo}.} \bibinfo{year}{2021}\natexlab{}.
\newblock \showarticletitle{Procedural generation of dungeons’ maps and
  locked-door missions through an evolutionary algorithm validated with
  players}.
\newblock \bibinfo{journal}{\emph{Expert Systems with Applications}}
  \bibinfo{volume}{180} (\bibinfo{year}{2021}), \bibinfo{pages}{115009}.
\newblock


\bibitem[\protect\citeauthoryear{Prager, Troost, Br{\"u}ggenj{\"u}rgen,
  Melh{\'a}rt, Yannakakis, and Preuss}{Prager et~al\mbox{.}}{2019}]%
        {prager2019facets}
\bibfield{author}{\bibinfo{person}{Raphael~Patrick Prager},
  \bibinfo{person}{Laura Troost}, \bibinfo{person}{Simeon
  Br{\"u}ggenj{\"u}rgen}, \bibinfo{person}{D{\'a}vid Melh{\'a}rt},
  \bibinfo{person}{Georgios~N. Yannakakis}, {and} \bibinfo{person}{Mike
  Preuss}.} \bibinfo{year}{2019}\natexlab{}.
\newblock \showarticletitle{An experiment on game facet combination}.
\newblock \bibinfo{journal}{\emph{IEEE Conference on Computatonal Intelligence
  and Games, CIG}}  \bibinfo{volume}{2019-August}.
\newblock
\showISBNx{9781728118840}
\showISSN{23254289}
\urldef\tempurl%
\url{https://doi.org/10.1109/CIG.2019.8848073}
\showDOI{\tempurl}


\bibitem[\protect\citeauthoryear{Smith, Padget, and Vidler}{Smith
  et~al\mbox{.}}{2018}]%
        {ref:smith2018graph}
\bibfield{author}{\bibinfo{person}{Thomas Smith}, \bibinfo{person}{Julian
  Padget}, {and} \bibinfo{person}{Andrew Vidler}.}
  \bibinfo{year}{2018}\natexlab{}.
\newblock \showarticletitle{Graph-based generation of action-adventure dungeon
  levels using answer set programming}. In
  \bibinfo{booktitle}{\emph{Proceedings of the 13th International Conference on
  the Foundations of Digital Games}}. ACM, \bibinfo{pages}{52}.
\newblock


\bibitem[\protect\citeauthoryear{Summerville, Mari{\~n}o, Snodgrass,
  Onta{\~n}{\'o}n, and Lelis}{Summerville et~al\mbox{.}}{2017}]%
        {ref:summerville2017understanding}
\bibfield{author}{\bibinfo{person}{Adam Summerville},
  \bibinfo{person}{Julian~RH Mari{\~n}o}, \bibinfo{person}{Sam Snodgrass},
  \bibinfo{person}{Santiago Onta{\~n}{\'o}n}, {and} \bibinfo{person}{Levi~HS
  Lelis}.} \bibinfo{year}{2017}\natexlab{}.
\newblock \showarticletitle{Understanding mario: an evaluation of design
  metrics for platformers}. In \bibinfo{booktitle}{\emph{Proceedings of the
  12th international conference on the foundations of digital games}}.
  \bibinfo{pages}{1--10}.
\newblock


\bibitem[\protect\citeauthoryear{Togelius, Shaker, and Nelson}{Togelius
  et~al\mbox{.}}{2016}]%
        {togelius2016introduction}
\bibfield{author}{\bibinfo{person}{Julian Togelius}, \bibinfo{person}{Noor
  Shaker}, {and} \bibinfo{person}{Mark~J. Nelson}.}
  \bibinfo{year}{2016}\natexlab{}.
\newblock \showarticletitle{Introduction}.
\newblock In \bibinfo{booktitle}{\emph{Procedural Content Generation in Games:
  A Textbook and an Overview of Current Research}},
  \bibfield{editor}{\bibinfo{person}{Noor Shaker}, \bibinfo{person}{Julian
  Togelius}, {and} \bibinfo{person}{Mark~J. Nelson}} (Eds.).
  \bibinfo{publisher}{Springer}, \bibinfo{pages}{1--15}.
\newblock


\bibitem[\protect\citeauthoryear{van~der Linden, Lopes, and Bidarra}{van~der
  Linden et~al\mbox{.}}{2013}]%
        {ref:van2013designing}
\bibfield{author}{\bibinfo{person}{Roland van~der Linden},
  \bibinfo{person}{Ricardo Lopes}, {and} \bibinfo{person}{Rafael Bidarra}.}
  \bibinfo{year}{2013}\natexlab{}.
\newblock \showarticletitle{Designing procedurally generated levels}. In
  \bibinfo{booktitle}{\emph{Proceedings of the the second workshop on
  Artificial Intelligence in the Game Design Process}}.
\newblock


\bibitem[\protect\citeauthoryear{van~der Linden, Lopes, and Bidarra}{van~der
  Linden et~al\mbox{.}}{2014}]%
        {ref:van2014procedural}
\bibfield{author}{\bibinfo{person}{Roland van~der Linden},
  \bibinfo{person}{Ricardo Lopes}, {and} \bibinfo{person}{Rafael Bidarra}.}
  \bibinfo{year}{2014}\natexlab{}.
\newblock \showarticletitle{Procedural generation of dungeons}.
\newblock \bibinfo{journal}{\emph{IEEE Transactions on Computational
  Intelligence and AI in Games}} \bibinfo{volume}{6}, \bibinfo{number}{1}
  (\bibinfo{year}{2014}), \bibinfo{pages}{78--89}.
\newblock


\bibitem[\protect\citeauthoryear{Viana and dos Santos}{Viana and dos
  Santos}{2021}]%
        {ref:viana2021procedural}
\bibfield{author}{\bibinfo{person}{Breno M~F Viana} {and}
  \bibinfo{person}{Selan~R dos Santos}.} \bibinfo{year}{2021}\natexlab{}.
\newblock \showarticletitle{Procedural Dungeon Generation: A Survey}.
\newblock \bibinfo{journal}{\emph{Journal on Interactive Systems}}
  \bibinfo{volume}{12}, \bibinfo{number}{1} (\bibinfo{year}{2021}),
  \bibinfo{pages}{83--101}.
\newblock


\end{thebibliography}

\end{document}